\documentclass{article}
\usepackage{graphicx} %

\usepackage{amsmath} %
\usepackage{amssymb}  %
\usepackage{amsthm}  %
\usepackage{bm}
\usepackage{enumitem}

\usepackage{lipsum}
\usepackage{color}

\newtheorem{theorem}{Theorem}[section]
\newtheorem{proposition}[theorem]{Proposition}
\newtheorem{lemma}[theorem]{Lemma}

\newtheorem*{theorem*}{Theorem}

\theoremstyle{definition}
\newtheorem{definition}[theorem]{Definition}
\newtheorem{remark}[theorem]{Remark}

\numberwithin{equation}{section}

\newcommand{\N}{\mathbb{N}}

\newcommand{\R}{\mathbb{R}}

\newcommand{\dd}{\,\mathrm{d}}
\newcommand{\norm}[1]{\left\lVert #1 \right\rVert}

\newcommand{\relu}{\operatorname{ReLU}}

\newcommand{\Mcal}{\mathcal{M}}

\newcommand{\Fcvx}{\mathcal{F}^{\mathrm{cvx}}}
\newcommand{\Fscvx}{\mathcal{F}^{\mathrm{scvx}}}
\newcommand{\Fqcvx}{\mathcal{F}^{\mathrm{qcx}}}

\newcommand{\Fqcc}{\mathcal{F}^{\mathrm{ct}}}

\newcommand{\Nicnn}{\mathcal{N}^{\mathrm{ICNN}}}
\newcommand{\Nscvx}{\mathcal{N}^{\mathrm{IsCNN}}}
\newcommand{\Nmono}{\mathcal{N}^{\mathrm{Imono}}}
\newcommand{\Ncc}{\mathcal N^{\mathrm{IctNN}}}
\newcommand{\Niq}{\mathcal{N}^{\mathrm{IqCNN}}}

\usepackage{amssymb}
\usepackage{hyperref}

\usepackage{graphicx}
\usepackage{caption}
\usepackage{subcaption}
\usepackage{comment}
\usepackage{algorithm}
\usepackage{algpseudocode}
\usepackage{booktabs}
\usepackage{multirow}
\usepackage{multicol}
\usepackage{xcolor}
\usepackage{natbib}
\graphicspath{{./figures/}}

\title{Structure-Preserving Neural Approximation of Convex, Semiconvex, and Quasiconvex Functions}

\author
{
Yeachan Park\thanks{Sejong University, 209 Neungdong-ro, Gwangjin-gu, Seoul, 05006, Seoul, South Korea}
 \and
 Yeoneung Kim\thanks{Yonsei University, 50 Yonsei-ro, Seodaemun-gu, Seoul, 03722, Seoul, South Korea}
 \thanks{Corresponding author}
}

\date{}

\begin{document}

\maketitle

\begin{abstract}
Input convex neural networks provide a natural way to enforce convexity globally through architectural constraints. We study how this structure-preserving approximation framework extends to semiconvex, quasiconvex, and convex transformable functions. For $C^{1,1}$ convex functions on compact convex domains, we establish the sharp $W^{1,p}$ minimax rate $m^{-1/d}$ under max-affine approximation with $m$ affine pieces, and show that every max-affine function admits an exact bounded-width ICNN representation with complexity linear in $m$. The same $W^{1,p}$ approximation rate extends to semiconvex functions through quadratic lifting and to convex transformable functions through monotone composition. For general Lipschitz quasiconvex functions, we develop a quantitative finite-probe approximation theory based on monotone responses to affine projections and obtain a structure-preserving IqCNN with an $L^\infty$ approximation guarantee. We further show that these approximation results extend to ReLU-transferable smooth activations, including Softplus and ELU. Numerical experiments on PDE-based approximation problems compare the proposed architectures with parameter-matched MLPs and validation-tuned shape penalties, demonstrating global structural validity without penalty tuning while maintaining competitive approximation accuracy. Overall, the results identify recoverable convex latent structure as the key mechanism enabling gradient-level structure-preserving approximation beyond convexity.
\end{abstract}

\section{Introduction}
\label{sec:intro}

Convexity is a useful structural prior in neural approximation because it can be enforced globally through the network architecture. Input convex neural networks \citep[ICNNs;][]{amos2017input}, for example, guarantee convexity of the input--output map by constraining hidden-to-hidden weights to be nonnegative. Such guarantees are attractive in optimization, control, inverse problems, and related applications where violations of the prescribed geometry may matter independently of prediction error \citep{chen2019optimal,huang2021convex,makkuva2020optimal}.

An approximation theory for such models should address two questions simultaneously. First, how accurately can a target function be approximated while preserving its prescribed structure? Second, which structural assumptions are strong enough to control not only function values but also gradients? The second question is particularly relevant in optimization- and PDE-based settings, where \(W^{1,p}\) approximation provides substantially more information than value approximation alone.

This paper studies these questions for convex, semiconvex, quasiconvex, and convex transformable functions. The central distinction is whether the target admits a recoverable convex latent representation.

A \(\rho\)-semiconvex function \(s\) satisfies
\[
s(x)+\frac{\rho}{2}\|x\|^2
\quad\text{is convex},
\]
so convexity is recovered through a quadratic lift. A convex transformable function has the form $f=\phi\circ g$,
where \(g\) is convex and \(\phi\) is nondecreasing. Thus, both semiconvex and convex transformable functions retain a single convex latent function. General quasiconvexity is weaker: \(f\) is quasiconvex if
\[
f(\lambda x+(1-\lambda)y)
\leq
\max\{f(x),f(y)\},
\qquad \lambda\in[0,1],
\]
or equivalently if all of its sublevel sets are convex. In general, this condition does not provide a single convex latent representation.

Our results show that this structural distinction is reflected directly in the approximation theory. Convex, semiconvex, and convex transformable functions admit \(W^{1,p}\) approximation with the convex exponent, whereas general quasiconvex functions are treated through a finite-probe \(L^\infty\) approximation theory. Table~\ref{tab:approx-regimes} summarizes the corresponding regimes.

\begin{table}[t]
\centering
\caption{Structure-preserving approximation regimes considered in this paper.
Each neural-network result holds for ReLU and for ReLU-transferable
activations, including Softplus and ELU, with the same approximation order.
Here $\varepsilon$ denotes the target approximation error.}
\label{tab:approx-regimes}
\begin{tabular}{lllll}
\toprule
Function class
& Latent structure
& Structured model
& Norm
& Complexity \\
\midrule
Convex
& Convex
& Max-affine / ICNN
& $W^{1,p}$
& $O(\varepsilon^{-d})$ \\

Semiconvex
& Quadratic lift to convex
& IsCNN
& $W^{1,p}$
& $O(\varepsilon^{-d})$ \\

Convex transformable
& Monotone transform of convex
& IctNN
& $W^{1,p}$
& $O(\varepsilon^{-d})$ \\

General quasiconvex
& Family of affine probes
& IqCNN
& $L^\infty$
& $O(\varepsilon^{-(d+1)})$ \\
\bottomrule
\end{tabular}
\end{table}

For \(C^{1,1}\) convex functions on a compact convex domain \(K\subset\mathbb{R}^d\), we first study approximation by maxima of \(m\) affine functions. We prove the sharp minimax rate $m^{-1/d}$ in \(W^{1,p}(K)\). 
The upper bound is constructive and is obtained from tangent planes sampled on a finite covering of the domain, while the lower bound is derived from a quadratic benchmark and a quantization argument for its gradient. We further show that every max-affine function with $m$ pieces can be represented exactly by a width-one sequential ICNN of depth \(m-1\), with parameter count linear in $m$. The max-affine approximation theorem therefore yields an explicit ICNN upper bound of order \(O( \varepsilon^{-d})\).

For semiconvex functions, the convex approximation theory transfers directly through quadratic lifting. If $s(x)=g(x)-\frac{\rho}{2}\|x\|^2$ with \(g\) convex, then approximation of \(g\) transfers without loss to \(s\) in \(W^{1,p}\). This leads naturally to the input semiconvex neural network (IsCNN), consisting of an ICNN together with a fixed quadratic correction.

General quasiconvex functions require a different representation. A continuous quasiconvex function can be written as
$f(x)=\sup_{a\in\mathbb{S}^{d-1}} \phi_a(a^\top x), $
where each \(\phi_a\) is nondecreasing. We replace this infinite family by finitely many affine probes and approximate the associated one-dimensional monotone responses by monotone neural networks. This yields a quantitative finite-probe \(L^\infty\) approximation theorem and motivates the input quasiconvex neural network (IqCNN), whose hard-max form preserves quasiconvexity by construction.

Within the quasiconvex class, convex transformable functions provide a complementary single-latent representation. For $f=\phi\circ g,$ with \(g\) convex and \(\phi\) sufficiently regular and nondecreasing, approximation of the latent convex function propagates through the outer transformation. We thereby recover the convex \(W^{1,p}\) exponent and obtain an \(O(\varepsilon^{-d})\) approximation bound for the corresponding input convex transformable neural network (IctNN). Matching optimality is established for affine outer transformations.

These classes can be viewed along two related structural directions,
\[
\text{convex}
\longrightarrow
\text{semiconvex},
\]
and
\[
\text{convex}
\longrightarrow
\text{convex transformable}
\longrightarrow
\text{quasiconvex}.
\]
The first direction relaxes convexity through an explicit quadratic correction, while the second passes from a single convex latent function to progressively more general quasiconvex representations.

We also extend the theory beyond ReLU networks. For a class of convex, nondecreasing ReLU-transferable activations, including Softplus and ELU, the structure-preserving approximation results carry over with the same approximation orders.

Taken together, the results identify recoverable convex latent structure as the mechanism behind gradient-level approximation. Convex functions possess this structure directly, semiconvex functions recover it through quadratic lifting, and convex transformable functions retain it through monotone composition. General quasiconvex functions instead admit a representation through a family of affine probes, leading to a different approximation regime.

Numerical experiments in dimensions up to \(32\) compare the proposed architectures with parameter-matched multilayer perceptrons and validation-tuned shape penalties. The penalized models can substantially reduce sampled structural violations and can sometimes attain lower approximation error. The architecture-based models, however, enforce the prescribed geometry globally without tuning a shape-penalty parameter and achieve competitive accuracy together with greater stability in the high-dimensional convex transformable experiments.

The main contributions are as follows.

\begin{itemize}
    \item We establish a sharp \(W^{1,p}\) approximation theory for \(C^{1,1}\) convex functions, proving the minimax rate \(m^{-1/d}\) under max-affine approximation and showing that every max-affine function admits an exact bounded-width ICNN realization with complexity linear in the number of affine pieces.

    \item We extend the convex \(W^{1,p}\) approximation theory to broader classes with recoverable convex latent structure: semiconvex functions through quadratic lifting and convex transformable functions through monotone composition, leading to the corresponding IsCNN and IctNN architectures.

    \item For general Lipschitz quasiconvex functions, we develop a quantitative finite-probe approximation theory based on monotone responses to affine projections and introduce a structure-preserving IqCNN with an \(L^\infty\) approximation guarantee.

    \item We extend the approximation results to ReLU-transferable smooth activations, including Softplus and ELU, and provide PDE-based numerical experiments comparing the proposed structure-preserving architectures with parameter-matched MLPs and validation-tuned shape penalties. The experiments demonstrate global structural validity without penalty tuning while retaining competitive approximation accuracy.
\end{itemize}

The remainder of the paper is organized as follows. Section~\ref{sec:related} reviews related work. Section~\ref{sec:prelim} introduces the function classes and notation. Section~\ref{sec:arch} defines the neural architectures and proves exact max-affine realization by ICNNs. Section~\ref{sec:main} summarizes the main results. Sections~\ref{sec:cvx}, \ref{sec:scvx}, \ref{sec:qcx}, \ref{sec:cc} develop the approximation theory for convex, semiconvex, quasiconvex, and convex transformable functions, respectively. Section~\ref{sec:relutrans} treats ReLU-transferable activations, Section~\ref{sec:exp} presents the numerical experiments, and Section~\ref{sec:conclusion} concludes the paper.

\section{Related Work}
\label{sec:related}

\subsection{Input Convex Neural Networks and Max-Affine Approximation}
Input convex neural networks impose convexity of the input--output map through architectural constraints, most notably the nonnegativity of hidden-to-hidden weights \cite{amos2017input}. They have been applied to optimization-aware learning and control \citep{chen2019optimal}, optimal transport and generative modeling \citep{huang2021convex,makkuva2020optimal}, structured prediction, and energy-based modeling. Their approximation properties have also been studied, including the approximation of convex target functions by ICNNs \citep{huang2021convex}. ICNNs have further been used in a variety of control applications \citep{yang2021optimization,bunning2021input,lawrynczuk2022input}. Closely related are max-affine models, which form a canonical class for convex regression and piecewise-linear convex approximation \citep{magnani2009convex,balazs2015near}.

\subsection{Structure-Preserving Neural Architectures}

More broadly, neural architectures have been designed to preserve geometric and analytic properties beyond ordinary convexity. Examples include parameterized convex models \citep{kim2022parameterized}, polyconvex architectures for mechanics \citep{geuken2025novel}, and networks constrained to be Lipschitz continuous \citep{anil2019sorting,gouk2021regularisation} or monotone \citep{sill1997monotonic,daniels2010monotone}. These works encode structural priors directly in the network architecture, complementing classical universal approximation results for unconstrained neural networks \citep{cybenko1989approximation,hornik1991approximation,leshno1993multilayer,pinkus1999approximation,yarotsky2017error}.

\subsection{Convexity and Generalized Convexity in Optimization, Control and Engineering}

Convexity and related generalized convexity conditions are widely used in optimization, variational analysis, optimal control, dynamic programming, robotics, and computer vision.

Semiconvex functions arise in nonsmooth and nonconvex constrained optimization, where their generalized differential properties support optimality analysis and numerical methods \citep{mifflin1977semismooth}. In optimal control and Hamilton--Jacobi theory, the closely related notion of semiconcavity is a fundamental regularity property of value functions and viscosity solutions \citep{bardi1997optimal,cannarsa2004semiconcave}. Semiconvexity and semiconcavity have also been used in optimal control \citep{bonnet2022semiconcavity,federico2026partial} and robotics \citep{jiang2021hand}.

Quasiconvexity, understood through convexity of sublevel sets, provides another classical relaxation of convexity in mathematical programming. Quasiconvex optimization problems retain convex sublevel sets and can often be solved through sequences of convex feasibility problems or bisection procedures. This structure underlies both classical optimality theory for quasiconvex programs and modern disciplined quasiconvex programming frameworks \citep{linh2006optimality,eppstein2005quasiconvex,agrawal2020disciplined}. Quasiconvex objectives and constraints have been used in multiview geometry \citep{agarwal2008fast,hartley2007optimal}, control and remote estimation \citep{chakravorty2018sufficient}, and inverse optimization for applications such as production planning \citep{chan2019inverse}.

\section{Preliminaries}
\label{sec:prelim}

\subsection{Notations}
Throughout the paper, $K \subset \R^d$ is a compact convex domain with nonempty interior, and $0 \in \operatorname{int}(K)$. We write
\[
R_K := \sup_{x \in K} \|x\|.
\]
In addition, for fixed $\delta>0$, let $K_\delta$ denote the open $\delta$-neighborhood of $K$, defined by
\[
K_\delta := \{x\in\mathbb{R}^d:\operatorname{dist}(x,K)<\delta\}.
\]

\begin{definition}[$L^p$ and $W^{1,p}$ norms]
Let $1 \le p < \infty$. For a measurable function $f : K \to \R$, define
\[
\|f\|_{L^p(K)}
:=
\left(\int_K |f(x)|^p\,dx\right)^{1/p}.
\]
If $f$ is weakly differentiable, define
\[
\|f\|_{W^{1,p}(K)}
:=
\|f\|_{L^p(K)}+\|\nabla f\|_{L^p(K)}.
\]
\end{definition}

\begin{definition}[$M$-smoothness]
A differentiable function $f:K\to\R$ is called \emph{$M$-smooth} if its gradient is Lipschitz continuous with constant $M$, that is,
\[
\|\nabla f(x)-\nabla f(y)\|
\le
M\|x-y\|
\qquad
\forall x,y\in K.
\]
\end{definition}

\subsection{Convex-related function classes}
We study the hierarchy of convex-related function class. 
\begin{align*}
    \text{convex} \;&\subset\; \text{semiconvex}, \\
    \text{convex} \;&\subset\;
\text{convex transformable} \;\subset\; \text{quasiconvex}.
\end{align*}

First, we define a convex function and a class of convex functions with additional regularity. The normalization conditions $f(0)=0$ and $\nabla f(0)=0$ can be imposed without loss of generality by replacing $f$ with
\[
x\mapsto f(x)-f(0)-\langle \nabla f(0),x\rangle,
\]
which preserves convexity and smoothness. 
\begin{definition}[Convex function]
A function $g:K\to\R$ is called \emph{convex} if
\[
g(\lambda x+(1-\lambda)y)
\le
\lambda g(x)+(1-\lambda)g(y)
\qquad
\forall x,y\in K,\ \forall \lambda\in[0,1].
\]
\end{definition}

\begin{definition}[Convex function class]\label{def:convex}

For $M>0$,  we define a class of convex functions with regularity on open neighborhood $K_\delta$ as
\[
\mathcal{F}^{\mathrm{cvx}}(M):=
\left\{f|_K \middle|
\begin{array}{l}
f:K_\delta\to\mathbb{R}\text{ is convex and $M$-smooth},\\
f(0)=0,\quad \nabla f(0)=0
\end{array}
\right\}.
\]
\end{definition}
\begin{remark}
For example, if $K=[-R,R]$, $f(x)= x^4$, then on $K_\delta = (-R-\delta , R+\delta)$, we have $f''(x) = 12 x^2 \le 12 (R+\delta)^2$. Thus $f(x)$ belongs to $\Fcvx(M)$ with $M = 12 ( R + \delta)^2$. Although $f(x)=x^4$ itself is not globally $M$-smooth in $\R$, $f(x)$ belongs to $\Fcvx(M)$ on bounded domain $K$ and $K_\delta$.  
\end{remark}
Next, we define a semiconvex function and a class of semiconvex functions. 

\begin{remark}
If $g\in \Fcvx(M)$ is twice differentiable, then
\[
0 \preceq D^2 g(x)\preceq M I_d
\qquad\text{for all }x\in K.
\]
More generally, the same holds almost everywhere when the Hessian exists in the weak sense.
\end{remark}

\begin{definition}[Semiconvex function]
A function $s:K\to\R$ is called \emph{$\rho$-semiconvex} if
$s(x)+\frac{\rho}{2}\|x\|^2$ is convex on $K$.
\end{definition}

\begin{definition}[Semiconvex function class]
For $M>0$ and $\rho\ge0$, define
\[
\Fscvx(M, \rho):=\left\{s:K\to\R\;:\;s(x)+\frac{\rho}{2}\|x\|^2 \in \Fcvx(M)\right\}.
\]
\end{definition}

\begin{remark}
Every $s\in\Fscvx(M,\rho)$ admits the representation
\[s(x)=g(x)-\frac{\rho}{2}\|x\|^2\quad \text{for some } g\in\Fcvx(M).
\]
Thus semiconvexity is exactly convexity after quadratic lifting. For instance, $f(x)= \|x\|^4 - \|x\|^2$ and $g(x)=\sin(\|x\|^2)$ are semiconvex on any bounded domain but not convex.
\end{remark}

Next, we define a quasiconvex function and a class of quasiconvex functions.  
\begin{definition}[Quasiconvex function]
A function $f:K\to\R$ is called \emph{quasiconvex} if
\[f(\lambda x+(1-\lambda)y)\le\max\{f(x),f(y)\}\quad \forall x,y\in K,\ \forall \lambda\in[0,1].
\]
Equivalently, every sublevel set
\[
L_t(f):=\{x \in K : f(x)\le t\}
\]
is convex.
\end{definition}

\begin{definition}[Quasiconvex function class]
For $L>0$, define
\[
\Fqcvx(L):=\left\{f:K\to\R\;\middle|\;\begin{array}{l}f \text{ is quasiconvex on } K,\\f(0)=0, f \text{ is } L \text{-Lipschitz.}
\end{array}
\right\}.
\]
\end{definition}

Now, we introduce the notion of convex transformable functions. Some quasiconvex functions admit a useful representation as the composition of an increasing function with a convex function. convex transformable functions have been studied extensively in the generalized convexity literature \citep{schaible1981generalized,avriel2010generalized,khajavirad2014relaxations}.

\begin{definition}[Convex transformable function]
\label{def:cc}
A function $f:K\to\R$ is called \emph{convex transformable} if it can be written as
$f=\varphi\circ g, $
where $g$ is convex on $K$ and $\varphi:\R\to\R$ is monotone increasing.
\end{definition}

The following lemma ensures that the composition of an increasing function with a convex function is quasiconvex.
\begin{lemma}[Monotone composition]
\label{lem:monocomp}
Let $g:K\to\R$ be convex and let $\varphi:\R\to\R$ be increasing. Then $\varphi\circ g$ is quasiconvex. In particular, every convex transformable function is quasiconvex.
\end{lemma}

\begin{proof}
Since every convex function is quasiconvex,
\[ g(\lambda x+(1-\lambda)y) \le \max\{g(x),g(y)\}. \]
Applying the monotonicity of $\varphi$ gives
\[ \varphi(g(\lambda x+(1-\lambda)y)) \le \max\{\varphi(g(x)),\varphi(g(y))\}. \]
Hence $\varphi\circ g$ is quasiconvex.
\end{proof}

We next introduce a class of convex transformable functions with additional regularity. As discussed in Remark~\ref{rem:ct}, conditions $0\le \varphi'(t)\le 1,\; \operatorname{Lip}(\varphi')\le 1$  serve as normalization conditions rather than essential restrictions.

\begin{definition}[Convex transformable class]\label{def:ct}
For $L,M>0$, define
\[ \Fqcc(M):=\left\{ f=\varphi\circ u \;\middle| \; \begin{array}{l}f(0)=0,\quad \nabla f(0)=0, \\u \in \Fcvx(M),\\\varphi\in C^{1,1}(\mathbb{R})\text{ is increasing},\\0\le \varphi'(t)\le 1, \; \operatorname{Lip} (\varphi')\le 1,\end{array}\right\},\]
\end{definition}

\begin{remark}[Normalization of the outer transform]\label{rem:ct}
The conditions
\[
0\le \varphi'(t)\le 1,\qquad \operatorname{Lip}(\varphi')\le 1,
\]
in Definition~\ref{def:ct} are normalization conditions rather than essential restrictions on the class of convex transformable functions.
Indeed, suppose that $f=\varphi\circ u$ where $u$ is convex and $M_0$-smooth, and $\varphi\in C^{1,1}(\mathbb R)$
is increasing with
\[
A:=\|\varphi'\|_{L^\infty(\mathbb R)}<\infty,\qquad B:=\operatorname{Lip}(\varphi')<\infty.
\]
For any $a>0$, define
\[
u_a(x):=a\,u(x),
\qquad
\varphi_a(t):=\varphi(t/a).
\]
Then $f=\varphi_a\circ u_a$ and  $\varphi_a'(t)= \frac1a\varphi'(t/a)$, and hence
\[
\|\varphi_a'\|_{L^\infty(\mathbb R)}=\frac{A}{a},\qquad\operatorname{Lip}(\varphi_a')=\frac{B}{a^2}.
\]
Therefore, choosing $a\ge \max\{A,\sqrt{B}\}$ gives
\[
0\le \varphi_a'(t)\le 1,
\qquad
\operatorname{Lip}(\varphi_a')\le 1.
\]
The rescaling only changes the smoothness parameter of the latent convex function: if $u$ is $M_0$-smooth, then $u_a$ is $(aM_0)$-smooth. Consequently, any representation with bounded $\varphi'$ and Lipschitz
$\varphi'$ can be normalized to satisfy the conditions of Definition~\ref{def:ct}, at the expense of replacing the latent smoothness constant $M_0$ by $aM_0$. Thus, when the smoothness parameter $M$ is allowed to adjust accordingly, the bounds on $\varphi'$ are merely a convenient normalization of the
convex transformable representation.
\end{remark}

\begin{lemma}[Finite maxima preserve quasiconvexity]
\label{lem:maxquasi}
If $f_1,\dots,f_n:K\to\R$ are quasiconvex, then $\max_{1\le i\le n} f_i$ is quasiconvex.
\end{lemma}

\begin{proof}
For every $t\in\R$,
\[L_t\Bigl(\max_{1\le i\le n} f_i\Bigr)=\bigcap_{i=1}^n L_t(f_i),
\]
a finite intersection of convex sets, hence convex.
\end{proof}

A level-set characterization of Definition~\ref{def:cc}, together with a checkable second-order sufficient condition, is given in Section~\ref{sec:cc}.

\section{Input-convex-related neural network architectures}
\label{sec:arch}

This section introduces the neural-network architectures used throughout the paper: the ICNN for convex targets, its quadratic lift IsCNN for semiconvex targets, multi-branch monotone architectures IqCNN for general quasiconvex targets, and the compositional architecture IctNN for convex transformable targets. We also prove the exact realization theorem that transfers max-affine approximation rates to ICNNs.

\subsection{Input convex neural network (ICNN)}

We present the ICNN architecture \citep{amos2017input} for representing convex functions with neural networks.
\begin{definition}[Input convex neural network ]
An \emph{ICNN} is a ReLU network of the form
\[
z_{k+1}
=
\relu(W_k z_k + U_k x + b_k),
\qquad
k=0,\dots,L-1,
\]
with $z_0\equiv 0$, where each $W_k$ has entrywise nonnegative coefficients. The output is
\[
F(x)=c^\top z_L + a^\top x + \beta,
\qquad
c\ge0 \text{ entrywise.}
\]
We write $n(F)$ for the number of trainable parameters of $F$, and for $m\in\N$ we let $\Nicnn_m$ denote the class of ICNNs with $n(F)\le m$.
\end{definition}

\begin{lemma}[\citealp{amos2017input}]
\label{lem:icnn-convex}
Every ICNN is convex with respect to the input $x$.
\end{lemma}

\begin{proof}
The proof is by induction on the layers. Since $z_0\equiv 0$, each coordinate of $z_0$ is convex. If all coordinates of $z_k$ are convex, then each coordinate of $W_k z_k + U_k x+b_k$ is convex because $W_k\ge 0$ entrywise and $U_kx+b_k$ is affine. Since $\relu(r)=\max\{r,0\}$ is convex and nondecreasing, each coordinate of $z_{k+1}$ is convex. The final output is convex because it is a nonnegative linear combination of convex functions plus an affine term.
\end{proof}

\subsection{Exact realization of max-affine functions}

We present a class of max-affine functions. Since $\max(\cdot)$ can be realizable by $\relu(\cdot)$ functions (Lemma~\ref{lem:max}), max-affine functions are representable exactly by ICNNs (Proposition~\ref{prop:maxaffine_icnn}).

\begin{definition}[Max-affine class]
For $m\in\N$, let
\[\Mcal_m:=\left\{h(x)=\max_{1\le i\le m}\{a_i^\top x+b_i\}\;:\;h(0)=0 \right\}. \]
\end{definition}

\begin{lemma}\label{lem:max}
For all \(u,v\in\R\),
\[\max\{u,v\}=\relu(u-v)+v.\]
\end{lemma}

\begin{proof}
If \(u\ge v\), the right-hand side equals \(u-v+v=u\). If \(u<v\), it equals \(0+v=v\).
\end{proof}

\begin{proposition}[Exact max-affine realization]\label{prop:maxaffine_icnn}
Let
\[h(x)=\max_{1\le i\le m}\{a_i^\top x+b_i\}.\]
Then $h$ is represented exactly by an ICNN of hidden width $1$ and depth $m-1$. In particular, the number of trainable parameters is $(d+1)m$.
\end{proposition}

\begin{proof}
The proof is presented in Section~\ref{proof:prop:maxaffine_icnn}.
\end{proof}

This exact realization result is essential throughout the paper. If a max-affine approximation theorem is proved, it immediately becomes an ICNN approximation theorem.

\subsection{Input semiconvex neural network (IsCNN)}
We present the IsCNN architecture for representing semiconvex functions with neural networks.

\begin{definition}[Input semiconvex neural network]
For $m\in\N$ and $\rho\ge0$, define
\[
\Nscvx(\rho):=\left\{F(x)-\frac{\rho}{2}\|x\|^2\;:\;F\in\Nicnn\right\}.
\]
\end{definition}

By construction, every element of $\Nscvx_m(\rho)$ is $\rho$-semiconvex: the quadratic term is fixed, while the trainable component is represented by an ICNN. Similar to ICNNs, semiconvex functions can also be represented using lifted max-affine functions, which in turn can be represented by input semiconvex neural networks.

\begin{definition}[Semiconvex lifted max-affine class]
For $m\in\N$ and $\rho\ge0$, define
\[\Mcal_m^{\mathrm{scvx}}(\rho):=\left\{h(x)-\frac{\rho}{2}\|x\|^2\;:\;h\in\Mcal_m\right\}.
\]
\end{definition}

\subsection{Monotone networks and input quasiconvex neural networks}

We present the IqCNN architecture (Definition~\ref{def:IqCNN}) for representing quasiconvex functions with neural networks. Since general continuous quasiconvex functions can be represented as the maximum of compositions of monotone functions with convex functions (Lemma~\ref{lem:quasi_repr}), we represent general quasiconvex functions using monotone networks and ICNNs.
The design of an architecture for general quasiconvex targets rests on the following exact representation, which reformulates Proposition~2.2 of \citet{xu2001level}.

\begin{lemma}[Representation of continuous quasiconvex functions]\label{lem:quasi_repr}
Let \(f:K\to\R\) be continuous and quasiconvex. For each \(a\in\mathbb S^{d-1}\), define
\[
\phi_a(s) := \inf\{f(y):\ y\in K,\ a^\top y\ge s\}, \qquad s\in I_a:=\{a^\top x : x\in K\}.
\]
Then each \(\phi_a\) is monotone increasing, and
\[
f(x)=\sup_{a\in\mathbb S^{d-1}} \phi_a(a^\top x)
\qquad \forall x\in K.
\]
\end{lemma}

\begin{proof}
By taking $\varepsilon \to 0$ in Lemma~\ref{lem:quasi_repr_rate}, we get the desired result.
\end{proof}

\begin{remark}
The lemma shows that the exact structural description of a general continuous quasiconvex function is not a single latent convex map, nor a finite maximum of such maps, but an \emph{infinite} supremum of monotone responses over affine probes indexed by the sphere. This is the key structural distinction between the unrestricted quasiconvex class and the convex transformable class of Definition~\ref{def:cc}.
\end{remark}

Truncating the supremum to finitely many probes and replacing each monotone response by a monotone network yields a finite architecture. We first fix the one-dimensional monotone building block.

\begin{definition}[Monotone network class]
\label{def:mono}
Let $[a,b]\subset\R$ be a bounded interval and let $\sigma=\relu$. For $m\in\N$, define
\begin{align*}
&\Nmono_m:=  \\
&\left\{ x \mapsto \beta+\sum_{j=1}^{m} c_j\bigl(\sigma(x-t_{j-1})-\sigma(x-t_j)\bigr) : \beta\in\R, \; c_j\ge0, a=t_0<t_1<\dots<t_m=b \right\},    
\end{align*}
and $\Nmono:=\bigcup_m \Nmono_m$.
\end{definition}

\begin{remark}
Each summand $\sigma(x-t_{j-1})-\sigma(x-t_j)$ is nondecreasing, so the nonnegativity of the coefficients $c_j$ guarantees monotonicity of the network by construction. Every element of $\Nmono_m$ is a two-layer ReLU network with at most $2m$ hidden units, and conversely every nondecreasing continuous piecewise-linear function with knots $t_1,\dots,t_{m-1}$ belongs to $\Nmono_m$. Architectures of this type go back to \citet{sill1997monotonic}; see also \citet{daniels2010monotone}. The quantitative approximation property of $\Nmono_m$ is established in Lemma~\ref{lem:mono}.
\end{remark}

\begin{definition}[Input quasiconvex neural network (IqCNN)]\label{def:IqCNN}
\[\Niq:=\left\{x\mapsto\max_{1\le i\le n}\phi_i\bigl(F_i(x)\bigr)\;:\;n\in\N,\ F_i\in\Nicnn,\ \phi_i\in\Nmono\right\}.\]
\end{definition}

By Lemma~\ref{lem:monocomp}, each branch $\phi_i\circ F_i$ is quasiconvex, and by Lemma~\ref{lem:maxquasi} a finite maximum of quasiconvex functions is again quasiconvex, because its sublevel sets are finite intersections of convex sets. Hence \emph{every element of $\Niq$ is quasiconvex}: the max architecture is structure-preserving.

\subsection{Input convex transformable neural network (IctNN)}

We present the IctNN architecture (Definition~\ref{def:nct}) for representing convex transformable functions with neural networks. Certain quasiconvex functions admit a particularly simple representation: a single composition of a monotone network with an ICNN is sufficient. This structure allows such functions to be represented efficiently using neural networks.
We also introduce the transformable max-affine class (Definition~\ref{def:tma}).

\begin{definition}[Input convex transformable neural network (IctNN)]\label{def:nct}
\[\mathcal N^{\mathrm{IctNN}}:=\left\{\phi\circ F\;:\;F\in\Nicnn, \phi\in\Nmono\right\}.\]
\end{definition}
By Lemma~\ref{lem:monocomp}, every element of $N^{\mathrm{IctNN}}$ is quasiconvex.
\begin{definition}[Transformable max-affine class]\label{def:tma}
For $m\in\N$, define
\[\Mcal_m^{\mathrm{ct}} := \left\{ \varphi\circ h \;: \; h\in\Mcal_m,  \; \varphi\in C^{1,1}(\mathbb{R})\text{ is non-decreasing} \right\}. \]
\end{definition}

\subsection{Transfer to convex smooth activations}
So far, we have considered only $\relu$ activation functions. We now extend the analysis from $\relu$ to convex, $\relu$-like activations, thereby covering a more general class of activation functions.

\begin{definition}[$\relu$-transferable activation]
A function $\sigma:\mathbb R\to\mathbb R$ is called $\relu$-transferable if $\sigma : \mathbb{R} \to \mathbb{R}$ is $C^1$, convex and nondecreasing and  there exist positive numbers $\alpha_\tau,\beta_\tau>0$ and constants $\gamma_\tau\in\mathbb R$, indexed by $\tau\to\infty$, such that
$$
\rho_\tau(t) := \alpha_\tau \sigma(\beta_\tau t)+\gamma_\tau, \qquad  \rho_\tau(t)\longrightarrow \relu(t)
$$
locally uniformly on $\mathbb R$. Additionally, for almost every $t\neq 0$,
$ \rho_\tau'(t) \longrightarrow \mathbf 1_{\{t>0\}},$
and there exists $C_\sigma>0$, independent of $\tau$, such that
$0\le \rho_\tau'(t)\le C_\sigma \; \text{for a.e. }t\in\mathbb R. $
\end{definition}
Among widely used activation functions, the Softplus and ELU \citep{clevert2015fast} activation is $\relu$-transferable.
Our notion extends the softplus-type activations of \citet{huang2021convex}. Every sufficiently smooth softplus-type activation is ReLU-transferable, while our definition also permits activations that approach ReLU under more general positive rescalings such as ELU.

\begin{remark}[Softplus]
The Softplus activation
$\sigma(t)=\log(1+e^t)$ is $\relu$-transferable. Indeed, taking
$$ \alpha_\tau=\frac1\tau,\quad  \beta_\tau=\tau,\quad\gamma_\tau=0,$$
gives
$\rho_\tau(t)=\frac1\tau\log(1+e^{\tau t}),$ leading to $\rho_\tau(t)\to \relu(t)$ locally uniformly and
$\rho_\tau'(t)=\frac{1}{1+e^{-\tau t}}\longrightarrow\mathbf 1_{\{t>0\}}$
for every $t\neq0$, with
$0\le\rho_\tau'(t)\le1.$
\end{remark}

\begin{remark}[ELU]
The ELU activation
$$\sigma(t)=\begin{cases}
t, \quad & t \ge 0 \\
\alpha_0(e^t-1) \quad & t < 0,
\end{cases}$$ is $\relu$-transferable if $\alpha_0 = 1$. Indeed, taking
$$ \alpha_\tau=\frac1\tau,\quad  \beta_\tau=\tau,\quad\gamma_\tau=0,$$
gives
$$\rho_\tau(t)=\begin{cases}
t, \quad & t \ge 0 \\
\frac{\alpha_0}{\tau} (e^{\tau t}-1) \quad & t < 0,
\end{cases}$$ 
leading to $\rho_\tau(t)\to \relu(t)$ locally uniformly. 
And we have 
$$\rho_\tau'(t)=
\begin{cases}
1, \quad & t \ge 0 \\
\alpha_0 e^{\tau t} \quad & t < 0,
\end{cases} \longrightarrow\mathbf 1_{\{t>0\}}$$
for every $t\neq0$, with
$0\le\rho_\tau'(t)\le \alpha_0 = 1. $
\end{remark}

Thus the proposition applies to an ICNN using the fixed standard Softplus/ELU activation; the parameter $\alpha_\tau, \beta_\tau$ are absorbed into the network weights rather than into the activation itself.

We use $\sigma$-ICNN, $\sigma$-IsCNN, $\sigma$-IqCNN, and $\sigma$-IctNN to denote ICNN-family networks that use a $\relu$-transferable activation function $\sigma$.

\subsection{Proof of Proposition~\ref{prop:maxaffine_icnn}}
\label{proof:prop:maxaffine_icnn}
\begin{proof}
Write
\[
u_i(x):=a_i^\top x+b_i,\qquad i=1,\dots,m,
\]
and define the running maxima
\[s_1(x):=u_1(x), \quad s_k(x):=\max\{u_1(x),\dots,u_k(x)\}
\quad(k=2,\dots,m).\]
By Lemma~\ref{lem:max},
\[
s_{k+1}(x)= \max\{ s_k(x), u_{k+1}(x) \} = \relu\!\big(s_k(x)-u_{k+1}(x)\big)+u_{k+1}(x),\]
which admits a scalar hidden recursion.
Let \(z_k(x):=\relu(s_k(x)-u_{k+1}(x))\) for \(k=1,\dots,m-1\). Then
\[z_1(x)=\relu(u_1(x)-u_2(x)),\]
which is realized by the first layer with
\[W_0=[0],\qquad U_0=(a_1-a_2)^\top,\qquad b_0=b_1-b_2.
\]
Assume inductively that the $k$-th hidden unit computes \(z_k(x)=\relu(s_k(x)-u_{k+1}(x))\). Since $s_k\ge u_{k+1}$ implies $z_k=s_k-u_{k+1}$ and $s_k<u_{k+1}$ implies $z_k=0$, in both cases
\[s_{k+1}(x)=z_k(x)+u_{k+1}(x),
\]
so
\[
z_{k+1}(x)=\relu(s_{k+1}(x)-u_{k+2}(x))=\relu\bigl(z_k(x)+u_{k+1}(x)-u_{k+2}(x)\bigr).
\]
This is realized by the layer
\[W_k=[1]\ge 0,\qquad U_k=(a_{k+1}-a_{k+2})^\top,\qquad b_k=b_{k+1}-b_{k+2}.\]
At the final layer,
\[z_{m-1}(x)=\relu(s_{m-1}(x)-u_m(x)),\qquad h(x)=s_m(x)=z_{m-1}(x)+u_m(x),
\]
so the exact output is
\[
F(x)=1\cdot z_{m-1}(x)+a_m^\top x+b_m = h(x),
\]
with nonnegative output weight $c=[1]$. All hidden-to-hidden weights equal $0$ or $1$ and are in particular nonnegative, so $F$ is a valid ICNN. The width is $1$, the depth is $m-1$, and each layer uses $d$ scalar coefficients, for a total of $(d+1)m$ trainable parameters.
\end{proof}

\section{Summary of main results}
\label{sec:main}

We now summarize the main results of the paper. 
\subsection{Structure-preserving network design}

\begin{theorem}[Structure-preserving neural-network classes]
\label{thm:nn-design}
The following hold:
\begin{enumerate}[label=(\roman*)]
    \item every $F\in\Nicnn$ is convex, so $\Nicnn$ is a structure-preserving approximation class for $\Fcvx(M)$;
    \item every $F\in\Nscvx(\rho)$ is $\rho$-semiconvex, so $\Nscvx(\rho)$ is a structure-preserving approximation class for $\Fscvx(M,\rho)$;
    \item every $F\in\Ncc$ is convex transformable, so $\Ncc(M)$ is a structure-preserving approximation class for $\Fqcc(M)$;
    \item every $F\in\Niq$ is quasiconvex, so the $\Niq$ is a structure-preserving approximation class for the general quasiconvex family.
    \item Let $\sigma$ be $\relu$-transferable activation. Then $\sigma$-ICNN, $\sigma$-IsCNN, $\sigma$-IqCNN, and $\sigma$-IctNN are structure-preserving. 
\end{enumerate}
\end{theorem}

\begin{proof}
Item (i) is Lemma~\ref{lem:icnn-convex}. Item (ii) follows since $F(x)-\frac{\rho}{2}\|x\|^2+\frac{\rho}{2}\|x\|^2=F(x)$ is convex. Item (iii) is Lemma~\ref{lem:monocomp}. Item (iv) follows by combining Lemma~\ref{lem:monocomp} (each branch $\phi_i\circ F_i$ is quasiconvex) with Lemma~\ref{lem:maxquasi} (finite maxima preserve quasiconvexity). 
Item (v) follows by Theorem \ref{thm:trans_icnn} and Theorem \ref{thm:iqcnn_tranfer}. 
\end{proof}

\begin{remark}
Each target family admits a network model with matching latent structure: semiconvexity is handled by quadratic lifting, while convex transformableity is handled by monotone composition over a convex latent variable. In particular, the architectures never leave the target geometry during training, which is the source of the out-of-distribution robustness observed in Section~\ref{sec:exp}.
\end{remark}

\subsection{Approximation guarantees}

We present approximation guarantees for structure-preserving approximations of convex, semiconvex, quasiconvex, and convex transformable functions. For max-affine approximations of convex functions, we establish a minimax approximation bound. For ICNN, IsCNN, IqCNN, and IcNN approximations, we provide error bounds. Next, we extend the approximation results to $\relu$-transferable networks when the case $p=\infty$ is excluded.

\begin{theorem*}[Max-affine approximation of convex function, Theorem \ref{thm:icnn_maxaffine}]
Let $1\le p<\infty$ and $M>0$. There exist constants $c_{\mathrm{cvx}}, C_{\mathrm{cvx}}>0$, depending only on $d,p,$ and $K$, such that
\begin{align*}
 c_{\mathrm{cvx}} M m^{-1/d} \le    \sup_{g\in\Fcvx(M)} \inf_{h\in\Mcal_m}  \| g - h \|_{w^{1,p}(K)}  \le C_{\mathrm{cvx}} M m^{-1/d} \qquad
\forall m\ge1,
\end{align*}
\end{theorem*}

\begin{theorem*}[ICNN approximation of convex functions, Theorem \ref{thm:icnn}]
For every $g\in\Fcvx(M)$ and every $\varepsilon>0$, there exists an $F \in \Nicnn$ such that
\[
\|g-F\|_{W^{1,p}(K)}\le \varepsilon,
\qquad
n(F)\le (d+1) \left\lceil \left(\frac{C_{\mathrm{cvx}}M}{\varepsilon}\right)^d \right\rceil.
\]
\end{theorem*}

\begin{theorem*}[IsCNN approximation of semiconvex functions, Theorem \ref{thm:semicvx_net}]
For every $g\in\Fscvx(M,\rho)$ and every $\varepsilon>0$ there exist $F\in\Nscvx(\rho)$ such that
\[\|g-F\|_{W^{1,p}(K)}\le \varepsilon, \qquad n(F) \le (d+1) \left\lceil \left(\frac{C_{\mathrm{cvx}}M}{\varepsilon}\right)^d \right\rceil.\]
\end{theorem*}

\begin{theorem*}[IqCNN approximation of quasiconvex functions, Theorem \ref{thm:iqcnn_approx}]
Let $g \in \Fqcvx(L)$. Then for every $\varepsilon>0$ there exists $F\in\Niq$ such that
\[\| g -  F \|_{L^{\infty}(K)}  \le \varepsilon,\quad n(F) = O\bigl(\varepsilon^{-(d+1)}\bigr). \]
\end{theorem*}

\begin{theorem*}[IctNN approximation of convex transformable functions, Theorem \ref{thm:IctNN}]
Let $1\le p<\infty$. Then for every $g\in \Fqcc(M)$ and  $\varepsilon>0$ there exists $F\in \mathcal N^{\mathrm{IctNN}}$
such that
\[ \|g -F\|_{W^{1,p}(K)}\le \varepsilon, \quad n(F)=O(\varepsilon^{-d}).\]
\end{theorem*}

\begin{theorem*}[Transfer principle for $\relu$-transferable activations, Theorem \ref{thm:trans_icnn}, Theorem \ref{thm:iqcnn_tranfer}]
Let $1\le p<\infty$ and  $\sigma$ be $\relu$-transferable activation. 
Then the followings hold:
\begin{enumerate}
\item For given  $\relu$-ICNN, it can be approximated in $W^{1,p}(K)$  by $\sigma$-ICNNs preserving convexity.  
\item For given  $\relu$-IqCNN, it can be approximated uniformly on $K$ by $\sigma$-IqCNNs preserving quasiconvexity.  
\item For given  $\relu$-IctNN can be approximated in $W^{1,p}(K)$  by $\sigma$-IctNNs preserving convex transformability.
\end{enumerate}
Moreover, the approximation rates are same as $\relu$ networks.
\end{theorem*}
Therefore, combining together, we get the following $\sigma$-ICNN family approximation results.

\begin{theorem}[$\sigma$-ICNN family approximation]\label{thm:sigmaapprox}
Let $1\le p<\infty$ and  $\sigma$ be $\relu$-transferable activation. 
Then the followings hold:
\begin{enumerate}
\item Let $g \in \Fcvx$. Then for every $\varepsilon >0$, there exists an $\sigma$-ICNN $F$ such that 
\[ \|g-F\|_{W^{1,p}(K)}\le \varepsilon, \quad n(F)\le (d+1) \left\lceil \left(\frac{C_{\mathrm{cvx}}M}{\varepsilon}\right)^d \right\rceil.\]
\item Let $g\in\Fscvx(M,\rho)$. Then for  every $\varepsilon>0$, there exist $\sigma$-IsCNN $F$ such that
\[\|g-F\|_{W^{1,p}(K)}\le \varepsilon, \qquad n(F) \le (d+1) \left\lceil \left(\frac{C_{\mathrm{cvx}}M}{\varepsilon}\right)^d \right\rceil.\]
\item Let $g \in \Fqcvx(L)$. Then for every $\varepsilon>0$ there exists $\sigma$-IqCNN $F$ such that
\[\| g -  F \|_{L^{\infty}(K)}  \le \varepsilon,\quad n(F) = O\bigl(\varepsilon^{-(d+1)}\bigr). \]
\item Let $g\in \Fqcc(M)$. Then for every  $\varepsilon>0$ there exists $\sigma$-IctNN $F$
such that
\[ \|g -F\|_{W^{1,p}(K)}\le \varepsilon, \quad n(F)=O(\varepsilon^{-d}).\]
\end{enumerate}
\end{theorem}

\section{ICNN approximation of convex functions}
\label{sec:cvx}

This section proves Theorem~\ref{thm:icnn}. The upper bound is constructive and based on tangent-plane envelopes over a finite net; the lower bound is obtained from a single quadratic benchmark together with a quantization estimate for piecewise-affine gradients.

\subsection{Preliminary bounds}

\begin{lemma}[Growth bounds]
\label{lem:growth}
Let $g\in \Fcvx(M)$. Then for every $x\in K$,
\[
0\le g(x)\le \frac{M}{2}\norm{x}^2, \quad \norm{\nabla g(x)}\le M\norm{x}\le M R_K.
\]
Hence $g$ is $L$-Lipschitz with $L \le M R_K$. 
\end{lemma}

\begin{proof}
Since $\nabla g(0)=0$ and $g$ is $M$-smooth,
\[
\norm{\nabla g(tx)}=\norm{\nabla g(tx)-\nabla g(0)}\le M t\norm{x} \quad (t\in[0,1]).
\]
Integrating along the segment $t\mapsto tx$,
\[
g(x)=g(x)-g(0)=\int_0^1 \nabla g(tx)\cdot x\,\dd t \le \int_0^1 M t \norm{x}^2\,\dd t = \frac{M}{2}\norm{x}^2.
\]
Because $g$ is convex and $\nabla g(0)=0$, the point $0$ is a global minimizer, so $g(x)\ge g(0)=0$. The gradient bound follows from the Lipschitz estimate with $t=1$.
\end{proof}

\begin{definition}[Bregman divergence]
For a differentiable convex function \(g : K \to \R\), define
\[
D_g(x,y):=g(x)-g(y)-\nabla g(y)\cdot (x-y).
\]
\end{definition}

\begin{lemma}[Bregman bounds for smooth convex functions]\label{lem:bregman}
Let $g\in\mathcal{F}^{\mathrm{cvx}}(M)$ and set
\[
N_\delta:=\left\lceil\frac{\operatorname{diam}(K)}{\delta}\right\rceil+1.
\]
Then, for every $x,y\in K$,
\[
\frac{1}{2MN_\delta}\|\nabla g(x)-\nabla g(y)\|^2\le D_g(x,y)\le\frac{M}{2}\|x-y\|^2.
\]
\end{lemma}

\begin{proof}
The upper bound follows from the standard descent inequality.
Indeed, since $K$ is convex, the line segment joining $x$ and $y$
is contained in $K\subset K_\delta$. Hence $M$-smoothness gives
\[
g(x)\le g(y)+\nabla g(y)\cdot(x-y)+\frac{M}{2}\|x-y\|^2,
\]
and therefore
\[
D_g(x,y)\le\frac{M}{2}\|x-y\|^2.
\]
We now prove the lower bound. Fix $x,y\in K$, and write $N:=N_\delta.$
Subdivide the line segment from $y$ to $x$ into $N$ equal pieces: $x_j:=y+\frac{j}{N}(x-y)$ for $ j=0,\dots,N.$
Thus $x_0=y$, $x_N=x$, and, since $K$ is convex, $x_j\in K$ for all $j$
Moreover,
\begin{equation}
\|x_{j+1}-x_j\|=\frac{\|x-y\|}{N}\le\frac{\operatorname{diam}(K)}{N}<\delta.
\label{eq:breg1}    
\end{equation}
We first establish a local Bregman estimate for each consecutive
pair $(x_j,x_{j+1})$. Fix $j\in\{0,\ldots,N-1\}$ and define
\[
\psi_j(z):=g(z)-g(x_j)-\nabla g(x_j)\cdot(z-x_j),\quad z\in K_\delta.
\]
Since $g$ is convex, $\psi_j(z)\ge0$ for $ z \in K_\delta$.
Also, $\psi_j$ is $M$-smooth and
\[
\nabla\psi_j(z)= \nabla g(z)-\nabla g(x_j).
\]
Set
\[
u_j:=x_{j+1}-\frac1M\nabla\psi_j(x_{j+1}).
\]
By $M$-smoothness,
\[
\|u_j-x_{j+1}\|=\frac1M
\|\nabla g(x_{j+1})-\nabla g(x_j)\|\le
\|x_{j+1}-x_j\|<\delta,
\]
where the last inequality follows from (\ref{eq:breg1}).
Since $x_{j+1}\in K$, this implies
\begin{equation}
u_j\in K+\delta B=K_\delta.
\label{eq:breg_3}   
\end{equation}
We may therefore apply the descent inequality to $\psi_j$
between $x_{j+1}$ and $u_j$:
\[
\begin{aligned}
\psi_j(u_j)
&\le
\psi_j(x_{j+1})
+\nabla\psi_j(x_{j+1})
   \cdot(u_j-x_{j+1})
+\frac{M}{2}\|u_j-x_{j+1}\|^2
\\
&=
\psi_j(x_{j+1})
-\frac1M\|\nabla\psi_j(x_{j+1})\|^2
+\frac1{2M}\|\nabla\psi_j(x_{j+1})\|^2
\\
&=
\psi_j(x_{j+1})
-\frac1{2M}\|\nabla\psi_j(x_{j+1})\|^2.
\end{aligned}
\]
Combining this with $\psi_j(u_j)\ge0$ from~(2), we obtain
\begin{equation}
D_g(x_{j+1},x_j)=\psi_j(x_{j+1})\ge\frac1{2M}\|\nabla g(x_{j+1})-\nabla g(x_j)\|^2.
\label{eq:breg_4}    
\end{equation}
It remains to combine these local estimates. Set $v:=\frac{x-y}{N},$
so that $x_{j+1}-x_j=v$. Then we have 
\[
\sum_{j=0}^{N-1}D_g(x_{j+1},x_j)=g(x)-g(y)- \sum_{j=0}^{N-1}\nabla g(x_j)\cdot v.
\]
On the other hand,
\[
D_g(x,y)= g(x)-g(y)-N\nabla g(y)\cdot v.
\]
Consequently,
\begin{equation}
D_g(x,y)-\sum_{j=0}^{N-1}D_g(x_{j+1},x_j)=\sum_{j=0}^{N-1}\bigl(\nabla g(x_j)-\nabla g(y)\bigr)\cdot v.  \label{eq:breg_5}
\end{equation}
For every $j\ge1$, monotonicity of the gradient of a convex
function gives
\[\bigl(\nabla g(x_j)-\nabla g(y)\bigr)\cdot(x_j-y)\ge0.
\]
Since $x_j-y=jv,$ we have
\[
\bigl(\nabla g(x_j)-\nabla g(y)\bigr)\cdot v \ge0.
\]
Thus (\ref{eq:breg_5}) yields
\begin{equation}
D_g(x,y)\ge \sum_{j=0}^{N-1}D_g(x_{j+1},x_j).
\label{eq:breg_6}    
\end{equation}
Using (\ref{eq:breg_4}) in (\ref{eq:breg_6}),
\[
D_g(x,y)\ge\frac1{2M}\sum_{j=0}^{N-1}\|\nabla g(x_{j+1})-\nabla g(x_j)\|^2.
\]
Finally, by Cauchy--Schwarz,
\[
\|\nabla g(x)-\nabla g(y)\|^2=\left\|\sum_{j=0}^{N-1}\bigl(\nabla g(x_{j+1})-\nabla g(x_j)\bigr)\right\|^2\le N\sum_{j=0}^{N-1}\|\nabla g(x_{j+1})-\nabla g(x_j)\|^2.
\]
Therefore,
\[
D_g(x,y)\ge\frac1{2MN}\|\nabla g(x)-\nabla g(y)\|^2.
\]
Since $N=N_\delta$, this proves the lower bound.
\end{proof}

\begin{lemma}[Covering number]\label{lem:covering_lem}
There exists a constant \(C_K > 0\), depending only on \(K\), such that for every \(h\in (0,1]\) there exists a finite set \(X_h=\{x_1,\dots,x_N\}\subset K\) with
\[
K\subset \bigcup_{j=1}^N B(x_j,h),
\qquad
N\le C_K h^{-d}.
\]
Concretely, we have $C_K \le (2 \cdot \mathrm{diam}(K) +1 )^d $
\end{lemma}

\begin{proof}
This is the standard covering-number estimate for bounded subsets of \(\R^d\): since \(K\) is bounded, it is contained in a ball of finite radius and is therefore covered by at most \(C_K h^{-d}\) balls of radius \(h\).
\end{proof}

\subsection{Upper bound of max-affine approximation}

\begin{proposition}[Upper bound]\label{prop:cvx_upper_bound}
There exists a constant \(C>0\), depending only on \(d,p,K, \delta \), such that for every \(M>0\), every \(m\ge 2\), and every \(g\in \Fcvx(M)\), there exists \(h_m\in \Mcal_m\) satisfying
\[\norm{g-h_m}_{W^{1,p}(K)}\le C M m^{-1/d}. \]
Moreover, $h_m$ can be chosen so that $0\le h_m\le g$ on $K$, each slope of $h_m$ is a gradient value $\nabla g(x_j)$ at some point $x_j\in K$, and
\[ \norm{g-h_m}_{L^p(K)}\le C M m^{-2/d}. \]
In addition, we have
\[ |\nabla h_m(z) | \le C M. \]
\end{proposition}

\begin{proof}
Choose
\[r_m:=(2C_K)^{1/d} m^{-1/d}, \]
where $C_K$ is the covering number in Lemma~\ref{lem:covering_lem}. By Lemma~\ref{lem:covering_lem}, there exists $ X_{r_m} =  \{x_1,\dots,x_N\}\subset K$ such that $K\subset \bigcup_{j=1}^N B(x_j, r_m)$ with $N \le C_K r_m^{-d} \le m/2$. Let $X_m = \{ x_0 \} \cup  \{x_1,\dots,x_N\}$, where $x_0 = 0$.
For each $x_j\in X_m$, define the tangent plane
\[\ell_j(x):=g(x_j)+\nabla g(x_j)\cdot (x-x_j),\]
and set
\[ h_m(x):=\max_{0\le j\le N}\ell_j(x).
\]
Because \(g\) is convex, \(\ell_j\le g\) for all $j$, so \(h_m\le g\). Since $\ell_0\equiv 0$ (by $g(0)=0$ and $\nabla g(0)=0$), we have \(h_m(0)=0\) and $h_m\ge 0$, and thus \(h_m\in \Mcal_m\).

Fix \(x\in K\) and choose \(x_j\in X_m\) with \(\norm{x-x_j}\le r_m\). Since \(h_m(x)\ge \ell_j(x)\), Lemma~\ref{lem:bregman} gives
\[
0\le g(x)-h_m(x)\le g(x)-\ell_j(x)=D_g(x,x_j)\le \frac{M}{2}\norm{x-x_j}^2\le \frac{M}{2}r_m^2.
\]
Therefore
\begin{align}
   \norm{g-h_m}_{L^p(K)}\le \frac{M}{2}|K|^{1/p}r_m^2. \label{eq:hm_value}
\end{align}
Now fix a point \(z \in K\) at which \(h_m\) is differentiable, so that
\[
h_m(z) = \ell_{i_z}(z), \qquad \nabla h_m(z)=\nabla g(x_{i_z}),
\]
for some index $i_z \in \{0,\dots, N\}$. Then
\[
D_g(z,x_{i_z})=g(z)-\ell_{i_z}(z)=g(z)-h_m(z)\le \frac{M}{2}r_m^2,
\]
and the lower bound of Lemma~\ref{lem:bregman} yields
\[
\frac{1}{2M N_\delta }\norm{\nabla g(z)-\nabla g(x_{i_z})}^2\le \frac{M}{2}r_m^2,
\]
hence
\[
\norm{\nabla g(z)-\nabla h_m(z)}\le \sqrt{N_\delta}Mr_m \quad\text{for a.e. }z\in K,
\]
since $h_m$ is differentiable almost everywhere. Consequently,
\begin{align}
  \norm{\nabla g-\nabla h_m}_{L^p(K)}\le \sqrt{N_\delta} M |K|^{1/p} r_m.
  \label{eq:hm_gradient}
\end{align}
Combining \eqref{eq:hm_value} and \eqref{eq:hm_gradient},
\[
\norm{g-h_m}_{W^{1,p}(K)}\le \frac{M}{2}|K|^{1/p}r_m^2 + \sqrt{N_\delta} M|K|^{1/p}r_m \le C M m^{-1/d},
\]
for some $C=C(K,d,p, \delta)$. The additional properties in the statement are immediate from the construction, and the $L^p$ estimate \eqref{eq:hm_value} gives the improved value-level rate $m^{-2/d}$.
\end{proof}

\subsection{Lower bound of max-affine approximation}
\begin{lemma}\label{lem:ball_bound}
Let \(1\le p<\infty\). There exists a constant \( c_0= c_0(d,p)>0\) such that for every measurable set \(A\subset\R^d\) of finite measure and every \(a\in\R^d\),
\[\int_A \norm{x-a}^p dx \ge c_0 |A|^{1+p/d}.\]
In addition, 
\[ |\nabla h_m(z) | \le | \nabla g(z)| + M r_m \le M (R_K + r_m). \]
\end{lemma}

\begin{proof}
By translation invariance it is enough to take \(a=0\). Among all measurable sets of fixed measure, the integral of the radially increasing function \(x\mapsto \norm{x}^p\) is minimized by the centered Euclidean ball of the same volume. If \(|B_r|=|A|\), then
\[
\int_A \norm{x}^p\, dx
\ge
\int_{B_r}\norm{x}^p\, dx
=
\frac{d\,\omega_d}{d+p}r^{d+p},
\]
where $\omega_d = \pi^{d/2} / \Gamma(\tfrac d 2 +1)$ denotes the volume of the $d$-dimensional unit ball. Since \(|B_r|=\omega_d r^d\), this becomes
\[
\int_A \norm{x}^p\, dx
\ge
\frac{d\,\omega_d^{-p/d}}{d+p}|A|^{1+p/d},
\]
so that one may take $c_0(d,p)= \frac{d\,\omega_d^{-p/d}}{d+p}$.
\end{proof}

\begin{proposition}[Lower bound]\label{prop:lower_bound}
There exists a constant \(c = c(d,p,K)>0\) such that for every \(M>0\) and every \(m\ge 2\),
\[ \sup_{g\in\Fcvx(M)}\inf_{h\in\Mcal_m}\norm{g-h}_{W^{1,p}(K)} \ge c M m^{-1/d}. \]
\end{proposition}

\begin{proof}
Because \(0\in\operatorname{int}(K)\), there exists \(r_0>0\) such that $B_{r_0}\subset K$. Consider
\[
g_*(x):=\frac{M}{2}\norm{x}^2\in\Fcvx(M).
\]
Fix any \(h\in \Mcal_m\), say
\[
h(x)=\max_{1\le i\le m}\{a_i^\top x+b_i\}.
\]
Since \(h\) is piecewise affine, there exists a measurable partition of \(B_{r_0}\), up to a null set, into sets \(A_1,\dots,A_m\) such that
\[
\nabla h(x)=a_i
\qquad\text{for a.e. }x\in A_i.
\]
Therefore
\[
\norm{\nabla g_*-\nabla h}_{L^p(B_{r_0})}^p
\ge
\sum_{i=1}^m \int_{A_i} \norm{Mx-a_i}^p\,dx
=
M^p \sum_{i=1}^m \int_{A_i}\norm{x-a_i/M}^p\,dx.
\]
Applying Lemma~\ref{lem:ball_bound} to each \(A_i\),
\[
\norm{\nabla g_*-\nabla h}_{L^p(B_{r_0})}^p
\ge
c_0(d,p)\, M^p\sum_{i=1}^m |A_i|^{1+p/d}.
\]
Since \(\sum_{i=1}^m |A_i|=|B_{r_0}|\) and \(t\mapsto t^{1+p/d}\) is convex, Jensen's inequality gives
\[
\frac{1}{m}\sum_{i=1}^m |A_i|^{1+p/d}
\ge
\left(\frac{1}{m}\sum_{i=1}^m |A_i|\right)^{1+p/d} =    m^{-1-p/d}|B_{r_0}|^{1+p/d},
\]
hence
\[
\norm{\nabla g_*-\nabla h}_{L^p(K)}^p
\ge
c_0(d,p)\, |B_{r_0}|^{1+p/d}M^p m^{-p/d}.
\]
Taking \(p\)-th roots and using the fact that the \(W^{1,p}\)-norm dominates the gradient term yields
\[
\norm{g_*-h}_{W^{1,p}(K)}\ge c M m^{-1/d},
\]
with $c = c(d,p,K)$. Since \(h\in \Mcal_m\) was arbitrary, this proves the minimax lower bound.
\end{proof}

\subsection{ICNN approximation of convex functions}
We present the proof of the minimax bound for ICNN approximation of convex functions (Theorem~\ref{thm:icnn}). 

\begin{theorem}[Max-affine approximation of convex functions]
\label{thm:icnn_maxaffine}
Let $1\le p<\infty$ and $M>0$. There exist constants $c_{\mathrm{cvx}}, C_{\mathrm{cvx}}>0$, depending only on $d,p,$ and $K$, such that
\begin{align*}
 c_{\mathrm{cvx}} M m^{-1/d} \le    \sup_{g\in\Fcvx(M)} \inf_{h\in\Mcal_m}  \| g - h \|_{w^{1,p}(K)}  \le C_{\mathrm{cvx}} M m^{-1/d} \qquad
\forall m \ge 2,
\end{align*}
\end{theorem}

\begin{proof}
Propositions~\ref{prop:cvx_upper_bound} and~\ref{prop:lower_bound} give the two-sided bound. 
\end{proof}

\begin{theorem}[ICNN approximation of convex functions]
\label{thm:icnn}
For every $g\in\Fcvx(M)$ and every $\varepsilon>0$, there exists an $F \in \Nicnn$ such that
\[
\|g-F\|_{W^{1,p}(K)}\le \varepsilon,
\qquad
n(F)\le (d+1) \left\lceil \left(\frac{C_{\mathrm{cvx}}M}{\varepsilon}\right)^d \right\rceil.
\]
\end{theorem}
\begin{proof}
Given $\varepsilon>0$, choose
\[
m := \left\lceil\left(\frac{C_{\mathrm{cvx}}M}{\varepsilon}\right)^d\right\rceil,
\]
so that $C_{\mathrm{cvx}}M m^{-1/d}\le\varepsilon$. By Proposition~\ref{prop:cvx_upper_bound} there exists $h_m\in\Mcal_m$ within $\varepsilon$ of $g$ in $W^{1,p}(K)$, and by Proposition~\ref{prop:maxaffine_icnn} the function $h_m$ is realized exactly by an ICNN $F$ with $n(F) \le (d+1)m$.

\end{proof}

\section{Approximation of semiconvex functions}
\label{sec:scvx}

The semiconvex theory is obtained from the convex theory by an exact correspondence: quadratic lifting is a bijection between the target classes and between the approximation classes, and it preserves the $W^{1,p}$ distance.

\subsection{IsCNN approximation of semiconvex functions}
We present the main result of the minimax bound for IsCNN approximation of semiconvex functions (Theorem~\ref{thm:semicvx_net}). 

\begin{theorem}[Lifted max-affine approximation of semiconvex functions]
Let $1\le p<\infty$ and $M>0$. There exist constants $c_{\mathrm{scvx}}, C_{\mathrm{scvx}}>0$, depending only on $d,p,$ and $K$, such that
\begin{align*}
 c_{\mathrm{scvx}} M m^{-1/d} \le    \sup_{g\in\Fscvx(M,\rho)} \inf_{h\in\Mcal_m^{\mathrm{scvx}}(\rho)}  \| g - h \|_{W^{1,p}(K)}  \le C_{\mathrm{scvx}} M m^{-1/d} \qquad
\forall m\ge2,
\end{align*}
\end{theorem}

\begin{proof}
Let $g \in\Fscvx(M,\rho)$ and write $g = g_0 - \frac{\rho}{2}\|\cdot\|^2$ with $g_0 := g+\frac{\rho}{2}\|\cdot\|^2\in\Fcvx(M)$. By Theorem~\ref{thm:icnn_maxaffine}, 
\begin{align*}
 c_{\mathrm{cvx}} M m^{-1/d} \le    \sup_{g\in\Fcvx(M)} \inf_{h\in\Mcal_m}  \| g - h \|_{W^{1,p}(K)}  \le C_{\mathrm{cvx}} M m^{-1/d} \qquad
\forall m\ge2.
\end{align*}
For $h_0 \in \Mcal_m$, consider the bijection $h_0 \mapsto  h = h_0 -  \frac{\rho}{2}|\cdot\|^2$. Since $\| g_0 - h_0\|_{W^{1,p}(K)} = \| g_0 - h_0\|_{W^{1,p}(K)}$, we have 
\begin{align*}
 c_{\mathrm{scvx}} M m^{-1/d} \le    \sup_{g\in\Fscvx(M,\rho)} \inf_{h\in\Mcal_m^{\mathrm{scvx}}(\rho)}  \| g - h \|_{W^{1,p}(K)}  \le C_{\mathrm{scvx}} M m^{-1/d} \qquad
\forall m\ge2,
\end{align*}
with $c_{\mathrm{scvx}} = c_{\mathrm{cvx}}$ and $ C_{\mathrm{scvx}} = C_{\mathrm{cvx}}$.
\end{proof}

\begin{theorem}[IsCNN approximation of semiconvex functions]\label{thm:semicvx_net}
For every $g\in\Fscvx(M,\rho)$ and every $\varepsilon>0$ there exist $F\in\Nscvx(\rho)$ such that
\[
\|g-F\|_{W^{1,p}(K)}\le \varepsilon, \qquad n(F) \le (d+1) \left\lceil \left(\frac{C_{\mathrm{cvx}}M}{\varepsilon}\right)^d \right\rceil.
\]
\end{theorem}

\begin{proof}
Let $ g \in\Fscvx(M,\rho)$ and write $g = g_0 - \frac{\rho}{2}\|\cdot\|^2$ with $g_0 := g+\frac{\rho}{2}\|\cdot\|^2\in\Fcvx(M)$. By Theorem~\ref{thm:icnn}, there exists $F_0 \in\Nicnn$ such that
\[
\|g_0-F_0\|_{W^{1,p}(K)}\le \varepsilon, \qquad n(F_0)= (d+1) \left\lceil \left(\frac{C_{\mathrm{cvx}}M}{\varepsilon}\right)^d \right\rceil,
\]
Set $F := F_0 - \frac{\rho}{2}\|\cdot\|^2\in\Nscvx(\rho)$, then
\[
g - F = g_0 - F_0 \quad\text{on } K,
\]
and $\|g-F\|_{W^{1,p}(K)}=\|g_0-F_0\|_{W^{1,p}(K)}$. 
Since $n(F) = n(F_0)$, 
we have 
\[
\|g-F\|_{W^{1,p}(K)}\le \varepsilon, \qquad n(F)= (d+1) \left\lceil \left(\frac{C_{\mathrm{cvx}}M}{\varepsilon}\right)^d \right\rceil.
\]
\end{proof}

\section{Approximation of quasiconvex functions}
\label{sec:qcx}

We now turn to the unrestricted quasiconvex class. At this level, the latent convex structure used in the convex and semiconvex analyses is no longer available a priori. Instead, the correct structural description is the affine-probe supremal representation of Lemma~\ref{lem:quasi_repr}: a continuous quasiconvex function is recovered as the supremum of monotone one-dimensional responses applied to affine projections.

This representation is exact but inherently infinite-dimensional. Consequently, the natural finite-complexity approximation class is obtained by truncating the infinite supremum to finitely many affine probes. The results of this section should therefore be interpreted as a \emph{baseline} approximation theory for general quasiconvexity: they quantify what can be obtained at the fully general level, in terms of function values only, before any additional latent convex structure is imposed.

\subsection{Monotone network approximation of monotone functions}
\label{subsec:monotone_approx}

\begin{lemma}[Monotone $L^{\infty}$-approximation]\label{lem:mono-Lp}
Let $I=[a,b]$, and let $\phi:I\to\mathbb R$
be continuous and nondecreasing. Then, for every $m\in\mathbb N$, there exists
$h_m\in\mathcal N_m^{\mathrm{mono}}$
such that
\[
\|\phi-h_m\|_{L^\infty(I)} \le\frac{\operatorname{osc}_I\phi}{m}, \quad n(h_m) = 6m+1 = O(m), 
\]
where
\[
\operatorname{osc}_I\phi:= \max_{s\in I}\phi(s)-\min_{s\in I}\phi(s).
\]    
\end{lemma}
\begin{proof}
Set $\alpha:=\phi(a)$ and $
\beta:=\phi(b)$.
Since $\phi$ is nondecreasing, we have 
$\operatorname{osc}_I\phi=\beta-\alpha.$
If $\alpha=\beta$, then $\phi$ is constant, and the conclusion follows by taking
$h_m:=\phi$.

Assume that $\alpha<\beta$ . Define the equally spaced levels
\[
\lambda_j := \alpha+\frac{j}{m}(\beta-\alpha), \quad j=0,\dots,m.
\]
By continuity of $\phi$, for every $j$ there exists $t_j\in I$ such that
$\phi(t_j)=\lambda_j$.
The points $t_j$ may be chosen so that
\[
a=t_0<t_1<\cdots<t_m=b.
\]
Let $h_m$ be the continuous piecewise affine interpolation 
\[ 
h_m(x) = \alpha + \sum_{j=1}^m c_j \left( \sigma(x- t_{j-1}) - \sigma( x - t_j) \right).\]
Then $h_m$ 
satisfies
\[
h_m(t_j)=\phi(t_j)=\lambda_j, \quad j=0,\dots,m.
\]
with $n(h_m) = 6m + 1$.
On each interval $[t_{j-1},t_j]$, the monotonicity of $\phi$ gives
\[
\lambda_{j-1} \le \phi(s) \le \lambda_j.
\]
Since $h_m$ linearly interpolates between $\lambda_{j-1}$ and $\lambda_j$, one also has
\[ \lambda_{j-1} \le h_m(s) \le \lambda_j.
\]
Therefore,
\[
|\phi(s)-h_m(s)| \le \lambda_j-\lambda_{j-1}= \frac{\beta-\alpha}{m}
\]
for every $s\in[t_{j-1},t_j]$. Taking the supremum over $I$ yields
\[
|\phi-h_m|_{L^\infty(I)}\le\frac{\beta-\alpha}{m}=\frac{\operatorname{osc}_I\phi}{m}.
\]
Finally, the slopes of $h_m$ are
\[c_j\frac{\lambda_j-\lambda_{j-1}}{t_j-t_{j-1}}\ge0. \]
Hence $h_m\in\mathcal N_m^{\mathrm{mono}}$.
\end{proof}

\begin{lemma}[Monotone $W^{1,p}$-approximation]\label{lem:mono}
Let $f \in W^{2,p}(a,b)$, $1\le p\le\infty$, be nondecreasing. Then for every $m\in\N$ there exists $g_m \in \Nmono_m$ such that
\[\|f-g_m\|_{W^{1,p}(a,b)} \le \bigl(1+(b-a)\bigr)\, \frac{b-a}{m}\,\|f''\|_{L^p(a,b)}.\]
The derivative error contributes at order $(b-a)/m$ and the value error at order $(b-a)^2/m$, so the $W^{1,p}$ rate is governed by the derivative term. In particular, accuracy $\varepsilon$ in $W^{1,p}(a,b)$ is achieved with a monotone network of size $O(\varepsilon^{-1})$.
\end{lemma}

\begin{proof}
Since $f$ is nondecreasing,
\[
f'(x)\ge 0 \qquad \text{for a.e. }x\in(a,b).
\]
For $m\in\N$, let
\[
h:=\frac{b-a}{m},\qquad x_i:=a+ih,\quad i=0,\dots,m,
\]
and define
\[
s_i:=\frac{f(x_i)-f(x_{i-1})}{h}
=\frac1h\int_{x_{i-1}}^{x_i} f'(t)\,dt \ \ge\ 0.
\]
Set
\[
g_m(x):=f(a)+\sum_{i=1}^m s_i\Bigl(\sigma(x-x_{i-1})-\sigma(x-x_i)\Bigr)\in\Nmono_m.
\]
Then $g_m$ is nondecreasing and piecewise linear, and for a.e.\ $x\in(x_{i-1},x_i)$, $g_m'(x)=s_i$. Thus $g_m'$ is the piecewise-constant cell-average approximation of $f'$.

We first estimate the derivative error. Fix $i\in\{1,\dots,m\}$ and $x\in(x_{i-1},x_i)$. Since
\[
f'(x)-s_i
=
\frac1h\int_{x_{i-1}}^{x_i}\bigl(f'(x)-f'(t)\bigr)\,dt,
\qquad
f'(x)-f'(t)=\int_t^x f''(u)\,du,
\]
we obtain
\[
|f'(x)-s_i| \le \int_{x_{i-1}}^{x_i}|f''(u)|\,du.
\]
For $1\le p<\infty$, H\"older's inequality gives
\[
\int_{x_{i-1}}^{x_i}|f''(u)|\,du
\le h^{1-\frac1p}\|f''\|_{L^p(x_{i-1},x_i)},
\]
so that, integrating $|f'(x)-s_i|^p$ over $x\in(x_{i-1},x_i)$,
\[
\|f'-g_m'\|_{L^p(x_{i-1},x_i)}^p
\le h^p \|f''\|_{L^p(x_{i-1},x_i)}^p.
\]
Summing over $i=1,\dots,m$ yields
\[
\|f'-g_m'\|_{L^p(a,b)}
\le h\,\|f''\|_{L^p(a,b)},
\]
and for $p=\infty$ the same argument gives directly
\[
\|f'-g_m'\|_{L^\infty(a,b)} \le h\,\|f''\|_{L^\infty(a,b)}.
\]
Next, since $f(a)=g_m(a)$,
\[
f(x)-g_m(x)=\int_a^x \bigl(f'(t)-g_m'(t)\bigr)\,dt,
\]
hence
\[
\|f-g_m\|_{L^\infty(a,b)} \le \|f'-g_m'\|_{L^1(a,b)}
\le (b-a)^{1-\frac1p}\|f'-g_m'\|_{L^p(a,b)},
\]
by H\"older's inequality. Consequently,
\[
\|f-g_m\|_{L^p(a,b)}
\le (b-a)^{1/p}\|f-g_m\|_{L^\infty(a,b)}
\le (b-a)\|f'-g_m'\|_{L^p(a,b)}
\le (b-a)\,h\,\|f''\|_{L^p(a,b)}.
\]
Adding the derivative bound,
\[
\|f-g_m\|_{W^{1,p}(a,b)}
\le \bigl(1+(b-a)\bigr)\,h\,\|f''\|_{L^p(a,b)},
\]
with the same estimate for $p=\infty$. Recalling $h=(b-a)/m$ proves the claim.
\end{proof}

\subsection{Representation of quasiconvex functions}

The next lemma quantifies the truncation of the exact representation of Lemma~\ref{lem:quasi_repr} to finitely many probes, which is the structural backbone of the IqCNN architecture. The key point of the proof is the \emph{choice} of separating direction: an arbitrary separating hyperplane carries no control on the separation margin, but the direction obtained from the metric projection onto the sublevel set has margin at least the distance to the sublevel set, which the Lipschitz condition bounds from below uniformly.

\begin{lemma}
\label{lem:quasi_repr_rate}
Let \(f:K\to\R\) be continuous, quasiconvex, and \(L\)-Lipschitz.
Then for every \(\varepsilon>0\), there exist \(N\in\N\), vectors
\(a_1,\dots,a_N\in \mathbb S^{d-1}\), and monotone increasing functions
\(\phi_1,\dots,\phi_N\) such that
\[
0\;\le\; f(x) - \max_{1\le i\le N}\phi_i(a_i^\top x) \;\le\; \varepsilon
\qquad \forall x\in K,
\]
and moreover
\[
N \le \mathcal N\!\left(K,\frac{\varepsilon}{4L}\right),
\]
where \(\mathcal N(K,r)\) denotes the covering number of \(K\) by Euclidean balls of radius \(r\). In particular, if \(K\subset B(0,R)\), then
\[
N \le \left(1+\frac{8LR}{\varepsilon}\right)^d = O(\varepsilon^{-d}).
\]
\end{lemma}

\begin{proof}
Fix $\varepsilon>0$ and set $r:=\varepsilon/(4L)$. For each $a\in\mathbb S^{d-1}$, define $\phi_a$ as 
\[
\phi_a(s) := \inf\{f(y):\ y\in K,\ a^\top y\ge s\}, \qquad s\in I_a:=\{a^\top x : x\in K\}.
\]
Note that $\phi_a$ is well defined and monotone increasing, and $\phi_a(a^\top z)\le f(z)$ for all $z\in K$. This gives the upper inequality \(\max_i \phi_i(a_i^\top x)\le f(x)\) where $\phi_i := \phi_{a_i}$. \\
Fix $x\in K$ and set $\beta_x:=f(x)-\frac{\varepsilon}{2}.$
Define $L_\beta(f)$ as
\[L_\beta(f) := \{ x \in K  : f(x) \le \beta \}. \] 
We consider the following cases. \\
\textbf{Case 1: $L_{\beta_x}(f)=\emptyset$.}
Then $f(y)>\beta_x$ for all $y\in K$, so for any $a\in\mathbb S^{d-1}$ and $z\in K$,
\[
\phi_a(a^\top z)=\inf\{f(y):y\in K,\ a^\top y\ge a^\top z\}>\beta_x
=f(x)-\frac{\varepsilon}{2}.
\]
Fix any such \(a\) and call it \(a_x\). \\ 
\textbf{Case 2: \(L_{\beta_x}(f)\neq\emptyset\).}
Since \(f\) is continuous and quasiconvex, \(L_{\beta_x}(f)\) is a nonempty closed convex set, and \(x\notin L_{\beta_x}(f)\). For every $y\in L_{\beta_x}(f)$ we have $f(x)-f(y)\ge \varepsilon/2$ and the Lipschitz condition yields
\[
\|x-y\|\ge \frac{\varepsilon}{2L}, \qquad\text{hence}\qquad \operatorname{dist}\bigl(x,L_{\beta_x}(f)\bigr)\ge\frac{\varepsilon}{2L}.
\]
Let $P_x$ be the metric projection of $x$ onto $L_{\beta_x}(f)$ and define the unit vector
\[
a_x:=\frac{x-P_x}{\|x-P_x\|}.
\]
Convexity of $L_{\beta_x}(f)$ gives $(x-P_x)^\top(y-P_x)\le 0$ for all $y\in L_{\beta_x}(f)$, that is,
\[
a_x^\top y\le a_x^\top P_x=a_x^\top x-\|x-P_x\|
\qquad\forall y\in L_{\beta_x}(f).
\]
We define $ \alpha_x:=\sup_{y\in L_{\beta_x}(f)} a_x^\top y $, then
\[
\alpha_x \le  a_x^\top x-\operatorname{dist}\bigl(x,L_{\beta_x}(f)\bigr) \le a_x^\top x-\frac{\varepsilon}{2L}.
\]
Now let $z\in K$ with $\|z-x\| \le r=\varepsilon/(4L)$. Then
\[
a_x^\top z \ge a_x^\top x-\frac{\varepsilon}{4L} \ge \alpha_x+\frac{\varepsilon}{2L}-\frac{\varepsilon}{4L}= \alpha_x+\frac{\varepsilon}{4L}>\alpha_x.
\]
Hence every $z' \in K$ with 
\[a_x^\top z' \ge a_x^\top z > \alpha_x = \sup_{y\in L_{\beta_x}(f)} a_x^\top y \]
satisfies $z'\notin L_{\beta_x}(f)$, i.e., $f(z')>\beta_x$. Since the infimum is attained on the compact set $K$, it follows that
\[
\phi_{a_x}(a_x^\top z) = \inf\{f(z'):z'\in K,\ a^\top z' \ge a^\top z\} >\beta_x=f(x)-\frac{\varepsilon}{2}.
\]
Note that in both cases the conclusion
\[
\|z-x\|\le r \quad\Longrightarrow\quad \phi_{a_x}(a_x^\top z)>f(x)-\frac{\varepsilon}{2}
\]
holds with the uniform radius $r=\frac{\varepsilon}{4L}$, independent of $x$.
Finally, for such $z$, the Lipschitz condition gives 
\[f(x)\ge f(z)-L\|z-x\| \ge f(z)-\varepsilon/4,\] hence
\[
\phi_{a_x}(a_x^\top z)>f(x) - \frac{\varepsilon}{2} \ge f(z)-\frac{\varepsilon}{4}-\frac{\varepsilon}{2} >f(z)-\varepsilon.
\]
Since the radius $r$ is uniform in $z$, let $\{x_j\}_{j=1}^N\subset K$ be an $r$-net of $K$ with
\[
N\le \mathcal N\!\left(K,\frac{\varepsilon}{4L}\right),
\]
and set $a_j:=a_{x_j}$, $\phi_j:=\phi_{a_j}$. Fix $z\in K$ and choose $j$ with $\|z-x_j\|\le r$. Since
\[
f(z)-\varepsilon<\phi_j(a_j^\top z)\le f(z),
\]
hence
\[
f(z)-\varepsilon< \max_{1\le i\le N}\phi_i(a_i^\top z) \le f(z).
\]
Since $z\in K$ was arbitrary, the claimed two-sided bound holds on all of $K$. If $K\subset B(0,R)$, the standard volumetric bound yields
\[
\mathcal N(K,r)\le \left(1+\frac{2R}{r}\right)^d,
\]
and substituting \(r=\varepsilon/(4L)\) proves the result.
\end{proof}

\begin{remark}
The covering bound is unconditional: no separation, margin, or regularity assumption on the level sets is needed beyond continuity, quasiconvexity, and the Lipschitz condition. What fails without further structure is not the value-level rate but the \emph{gradient-level} theory, as discussed after Theorem~\ref{thm:iqcnn_approx}.
\end{remark}

\subsection{IqCNN approximation of quasiconvex functions}

\begin{theorem}[IqCNN approximation of quasiconvex functions]
\label{thm:iqcnn_approx}
Let $g \in \Fqcvx(L)$. Then for every $\varepsilon>0$ there exists $F\in\Niq$ such that
\[
\| g -  F \|_{L^{\infty}(K)}  \le \varepsilon,\quad n(F) = O\bigl(\varepsilon^{-(d+1)}\bigr).
\]
\end{theorem}

\begin{proof}
Fix $\varepsilon>0$ and set $\delta:=\frac{\varepsilon}{2}$, $\eta:=\frac{\varepsilon}{2}$.
By Lemma~\ref{lem:quasi_repr_rate}, there exist $N\in\N$, vectors
$a_1,\dots,a_N\in \mathbb S^{d-1}$, and monotone increasing functions
$\phi_1,\dots,\phi_N$ such that
\[
G(x):=\max_{1\le i\le N}\phi_i(a_i^\top x)
\]
satisfies
\[
|g(x)-G(x)| \le \delta \quad \forall x \in K \quad N \le \mathcal N\left(K,\frac{\delta}{4L}\right).
\]
In particular, $N = O(\delta^{-d}) = O(\varepsilon^{-d}).$
Furthermore,
\[
\operatorname{osc}_{I_i}\phi_i \le\ \operatorname{osc}_K g \le L  \cdot \mathrm{diam}(K).
\]
By Lemma~\ref{lem:mono-Lp}, there exists 
$h_m\in\mathcal N_m^{\mathrm{mono}}$ 
such that
\[
|\phi_i-h_{m_i}|_{L^\infty(I_i)} \le \frac{\operatorname{osc}_{I_i}\phi}{m_i}, \quad n(h_{m_i}) = O(m_i).
\]
Indeed, write
\[
\alpha_i:=\min_{I_i}\phi,\quad \beta_i:=\max_{I_i}\phi,\]
and define
\[
\lambda_j:=\alpha+\frac{j}{m}(\beta-\alpha),\quad j=0,\dots,m_i.\]
Choose $m_i := \left\lceil\frac{L \mathrm{diam}(K)}{\eta}\right\rceil.$ Then,
\[
|\phi_i-h_{m_i}|_{L^\infty(I_i)}\le\frac{\operatorname{osc}_I\phi_i}{m_i} \le \frac{L \cdot \mathrm{diam}(K)}{m_i} \le \eta, \quad n(h_{m_i}) = O(\eta^{-1}) = O(\varepsilon^{-1}).\]
Let $h_i = h_{m_i}$ and define
\[F(x):=\max_{1\le i\le N}h_i(a_i^\top x).
\]
Consequently, for every $x\in K$,
\[ |G(x)-F(x)|\le\max_{1\le i\le N}\left|\phi_i(a_i^\top x)-h_i(a_i^\top x)\right|\le\eta.
\]
Therefore,
\[
\|G-F\|_{L^\infty(K)}
\le
\eta.
\]
Combining the two uniform estimates gives
\[
\|g-F\|_{L^\infty(K)} \le \|g-G\|_{L^\infty(K)} + \|G-F\|_{L^\infty(K)} \le \delta+\eta = \varepsilon.
\]
Finally, we have 
\[
n(F)=O\left(N+\sum_{i=1}^N n(h_i)\right)=O\bigl(\varepsilon^{-(d+1)}\bigr).
\]
\end{proof}

\section{Approximation of convex transformable functions}
\label{sec:cc}

We now return to the structured quasiconvex subclass generated by compositions of convex latent functions and monotone maps. In this case, we can obtain a $W^{1,p}$ approximation result analogous to that in Section~\ref{sec:cvx}.

\subsection{Level-set characterization and a sufficient condition}

Definition~\ref{def:cc} describes convex transformableity through a factorization \(f=\varphi\circ g\). The following equivalent level-set formulation explains the terminology and shows that the class is intrinsic, i.e., independent of any particular factorization.

\begin{definition}[Level-set formulation]
\label{def:cc-levelset}
A continuous function $f:K\to\R$ is said to have \emph{convex transformable level structure} if there exists a continuous, increasing function $\psi:\R\to\R$ such that
\[
\lambda L_{\psi(r)}(f) + (1-\lambda) L_{\psi(s)}(f)
\;\subset\;
L_{\psi(\lambda r + (1-\lambda)s)}(f)
\qquad
\forall r,s\in\R,\ \forall\lambda\in[0,1],
\]
where $L_t(f):=\{x\in K: f(x)\le t\}$ and the addition on the left is in the Minkowski sense.
\end{definition}

\begin{lemma}[Level-set structure yields a convex factorization]
\label{lem:cc_repr}
Let $f:K\to\R$ be continuous with convex transformable level structure realized by a continuous, strictly increasing $\psi$ whose range covers $f(K)$. Then there exist a convex function $g:K\to\R$ and an increasing function $\varphi:\R\to\R$ such that
\[
f=\varphi\circ g \qquad\text{on }K.
\]
\end{lemma}

\begin{proof}
Define
\[
g(x):=\inf\{r\in\R:\ x\in L_{\psi(r)}(f)\}
=\inf\{r\in\R:\ f(x)\le \psi(r)\},
\]
which is finite for every $x\in K$ because the range of $\psi$ covers $f(K)$.

\smallskip
\noindent\emph{Step 1 : }
Fix $x\in K$ and let $r_n\downarrow g(x)$ with $f(x)\le\psi(r_n)$. Letting $n\to\infty$ and using the continuity of $\psi$ gives $f(x)\le\psi(g(x))$, that is, $x\in L_{\psi(g(x))}(f)$.

\smallskip
\noindent\emph{Step 2 : }
Fix $x,y\in K$ and $\lambda\in[0,1]$. By Step 1, $x\in L_{\psi(g(x))}(f)$ and $y\in L_{\psi(g(y))}(f)$. The Minkowski inclusion of Definition~\ref{def:cc-levelset} with $r=g(x)$, $s=g(y)$ yields
\[
\lambda x+(1-\lambda)y
\in
L_{\psi(\lambda g(x)+(1-\lambda)g(y))}(f),
\]
and hence, by the definition of $g$ as an infimum,
\[
g(\lambda x+(1-\lambda)y)\le \lambda g(x)+(1-\lambda)g(y).
\]
Thus $g$ is convex on $K$.

\smallskip
\noindent\emph{Step 3 : }
By Step 1, $f(x)\le \psi(g(x))$. Conversely, for every $r<g(x)$ we have $x\notin L_{\psi(r)}(f)$, i.e., $f(x)>\psi(r)$; letting $r\uparrow g(x)$ and using the continuity of $\psi$ gives $f(x)\ge \psi(g(x))$. Therefore $f=\psi\circ g$, and the claim follows with $\varphi:=\psi$.
\end{proof}

For $C^2$ functions there is a checkable differential sufficient (and, in the strictly monotone regime, necessary) condition. Conditions of this type are classical in the theory of convexifiable quasiconvex functions; we include a short self-contained proof.

\begin{lemma}[Second-order characterization]
\label{lem:suffcond_cc}
Let $f\in C^2(K)$ be quasiconvex. Then the following are equivalent:
\begin{enumerate}[label=(\roman*)]
\item there exists a continuous function $\mu:f(K)\to\R$ such that
\[
\nabla^2 f(x) + \mu(f(x))\, \nabla f(x)\nabla f(x)^\top \succeq 0
\qquad \forall x\in K;
\]
\item $f=\varphi\circ g$ on $K$ for some convex $g\in C^2(K)$ and some $C^2$, strictly increasing $\varphi$ with $\varphi'>0$ on $g(K)$.
\end{enumerate}
\end{lemma}

\begin{proof}
\emph{(i) $\Rightarrow$ (ii).}
Fix $t_0\in f(K)$ and define, on $f(K)$,
\[
\psi'(t):=\exp\!\left(\int_{t_0}^{t}\mu(s)\,ds\right)>0,
\qquad
\psi(t):=\int_{t_0}^{t}\psi'(s)\,ds,
\]
so that $\psi\in C^2$, $\psi$ is strictly increasing, and $\psi''=\mu\,\psi'$. Set $g:=\psi\circ f$. Then
\[
\nabla^2 g
=
\psi'(f)\,\nabla^2 f + \psi''(f)\,\nabla f\nabla f^\top
=
\psi'(f)\Bigl(\nabla^2 f + \mu(f)\,\nabla f\nabla f^\top\Bigr)
\succeq 0,
\]
since $\psi'>0$. Hence $g$ is convex, and $f=\varphi\circ g$ with $\varphi:=\psi^{-1}$, which is $C^2$ and strictly increasing with $\varphi'>0$.

\smallskip
\emph{(ii) $\Rightarrow$ (i).}
Let $\psi:=\varphi^{-1}$, so that $\psi\in C^2$, $\psi'>0$, and $g=\psi\circ f$. Convexity of $g$ gives
\[
0\preceq \nabla^2 g
=
\psi'(f)\,\nabla^2 f + \psi''(f)\,\nabla f\nabla f^\top.
\]
Dividing by $\psi'(f)>0$ yields
\[
\nabla^2 f + \frac{\psi''(f)}{\psi'(f)}\,\nabla f\nabla f^\top \succeq 0,
\]
so (i) holds with $\mu:=\psi''/\psi'$, which is continuous on $f(K)$.
\end{proof}

\subsection{Upper bound of convex transformable max-affine class}
\begin{proposition}[Upper bound]
\label{prop:cc_upper}
There exists a constant $C>0$, depending only on $d,p,K,M$, such that for every $M>0$, $m \ge 2$, $g \in \Fqcc(M)$, there exists $h_m \in \Mcal_{m}^{\mathrm{ct}}$
\begin{align*}
 \| g - h_m \|_{W^{1,p}(K)}  \le C M m^{-1/d} \quad \forall m\ge2,
\end{align*}
\end{proposition}

\begin{proof}
Write $g=\varphi\circ u$ where $u\in\Fcvx(M)$ and $\varphi$ is increasing with \[ 0\leq \varphi'\leq 1, \quad \operatorname{Lip}(\varphi')\leq 1. \]
By Proposition~\ref{prop:cvx_upper_bound}, there exists a constant $C_0$ and $q_m\in \Mcal_m$ such that
\[
\norm{u-q_m}_{W^{1,p}(K)}\le C_0 M m^{-1/d}, \quad \norm{u-q_m}_{L^p(K)}\le C_0 M m^{-2/d}.
\]
Let $g_m := \varphi \circ q_m \in \Mcal_{m}^{\mathrm{ct}}$.
Because $0\leq\varphi'\leq 1$ ,
\[
|g(x)-g_m(x)|= |\varphi(u(x))-\varphi(q_m(x))| \leq |u(x)-q_m(x)|,
\]
and 
\begin{equation}
\|g-h_m\|_{L^p(K)}\leq \|u-q_m\|_{L^p(K)}. 
\end{equation}
By the Sobolev chain rule,
\[
\nabla g=\varphi'(u)\nabla u, \quad \nabla h_m = \varphi'(q_m)\nabla q_m
\]
almost everywhere on $K$. Hence
\begin{align*}
\|\nabla g-\nabla h_m\|&=\|\varphi'(u)\nabla u-\varphi'(q_m)\nabla q_m\| \\& \leq|\varphi'(u)|\, \|\nabla u-\nabla q_m\|+ |\varphi'(u)-\varphi'(q_m)|\, \|\nabla q_m\|.
\end{align*}
Using
\[ |\varphi'|\leq 1, \quad |\varphi'(s)-\varphi'(t)|\leq |s-t|, \]
we have 
\[ \|\nabla g(x)-\nabla h_m(x)\| \le \|\nabla u-\nabla q_m\|+ |  u - q_m | \| \nabla q_m \|, \]
for almost every $x\in K$. Consequently, we have 
\[
\|\nabla g(x)-\nabla h_m(x)\|_{L^p(K)} \le \|\nabla u-\nabla q_m\|_{L^p(K)} + \|u - q_m \|_{L^p(K)} M R_K ,
\]
Adding the two estimates,
\begin{align*}
\norm{g-h_m}_{W^{1,p}(K)}
&\le \| u - q_m \|_{W^{1,p}(K)} + \|u - q_m \|_{L^p(K)} M R_K \\
&\le C_0 M m^{-1/d} +   C_0 M m^{-2/d} M R_K \le C m^{-1/d}.    
\end{align*}
where $C>0$ depends only on $d,p,K, M$.
\end{proof}

\subsection{IctNN approximation of convex transformable functions}

\begin{theorem}[IctNN approximation of convex transformable functions]
\label{thm:IctNN}
Let $1\le p<\infty$. Then for every $g\in \Fqcc(M)$ and  $\varepsilon>0$ there exists $F\in \mathcal N^{\mathrm{IctNN}}$
such that
\[ \|g-F\|_{W^{1,p}(K)}\le \varepsilon, \quad n(F)=O(\varepsilon^{-d}).\]
\end{theorem}

\begin{proof}
Let $g=\varphi\circ u\in\Fqcc(M)$, where $u \in \Fcvx(M)$ and $\varphi$ is increasing with $0 \le \varphi' \le 1$ and $\mathrm{Lip}(\varphi') \le 1$. 
By Proposition~\ref{prop:cc_upper}, for every $m\ge2$ there exists
$\widetilde g_m=\varphi\circ h_m\in\Mcal_m^{\mathrm{ct}}$,
with $h_m\in\Mcal_m$, such that
\begin{equation}
\label{eq:first-approximation}
\|g-\widetilde g_m\|_{W^{1,p}(K)}
\leq CMm^{-1/d}.
\end{equation}
Choose $m\geq\left(\frac{2CM}{\varepsilon}\right)^d.$
Then
$\|g-\widetilde g_m\|_{W^{1,p}(K)}
\leq \frac{\varepsilon}{2}.$
Write $h_m(x)=\max_{1\leq i\leq m}
\{a_i^\top x+b_i\}.$
Since $K$ is compact and $h_m$ is continuous, its range is a compact
interval
\[I_m=[\alpha_m,\beta_m],\quad\alpha_m:=\min_{x\in K}h_m(x), \quad \beta_m:=\max_{x\in K}h_m(x). \]
Set $R_m:=\beta_m-\alpha_m$.

Apply Lemma~\ref{lem:mono} to the restriction of $\varphi$ to $I_m$.
For every $q\in\mathbb N$, there exists $\psi_q\in\Nmono_q$ such that
\begin{equation}
\label{eq:mono-approximation}
\|\varphi-\psi_q\|_{W^{1,p}(I_m)}
\leq
(1+R_m)\frac{R_m}{q}
\|\varphi''\|_{L^p(I_m)}.
\end{equation}
Define $F_{m,q}:=\psi_q\circ h_m$.
By the definition of the IctNN, we have  $F_{m,q}\in\mathcal N^{\mathrm{IctNN}}$.
We now estimate the error between $\widetilde g_m$ and $F_{m,q}$.
For almost every $x\in K$, the Sobolev chain rule gives
\begin{align*}
 \nabla \widetilde g_m(x)&=\varphi'(h_m(x))\nabla h_m(x) \\   
\nabla F_{m,q}(x)&= \psi_q'(h_m(x))\nabla h_m(x).
\end{align*}
Thus
\begin{align*}
|\widetilde g_m(x)-F_{m,q}(x)| &\leq \|\varphi-\psi_q\|_{L^\infty(I_m)}  \\
\left| \nabla \widetilde g_m(x) -  \nabla F_{m,q}(x) \right| &\leq \|\varphi'-\psi_q'\|_{L^\infty(I_m)} |\nabla h_m(x)|.
\end{align*}
By Proposition~\ref{prop:cvx_upper_bound}, there exists $A>0$ depending only on $K,d,p$ such that 
$|\nabla h_m | \le A M$. 
Consequently,
\begin{align*}
\|\widetilde g_m-F_{m,q}\|_{W^{1,p}(K)}&\leq|K|^{1/p}\left(\|\varphi-\psi_q\|_{L^\infty(I_m)}+ A M  \|\varphi' - \psi_q'\|_{L^\infty(I_m)} \right)\\
&\leq C_1\|\varphi-\psi_q\|_{W^{1,\infty}(I_m)}.
\end{align*}
Constants may depend on the fixed parameters $d,p,K,M$, but not on $m$, $q$, or $\varepsilon$.
Alternatively, one may apply Lemma~\ref{lem:mono} with $p=\infty$ to
obtain
\[
\|\varphi-\psi_q\|_{W^{1,\infty}(I_m)}
\leq
(1+R_m)\frac{R_m}{q}
\|\varphi''\|_{L^\infty(I_m)}.
\]
Since the defining assumptions on $\Fqcc(M)$ give a uniform bound on
$\varphi''$ and the ranges $I_m$ are uniformly bounded, there exists
$C_2>0$ such that
\begin{equation*}
\|\widetilde g_m-F_{m,q}\|_{W^{1,p}(K)}
\leq \frac{C_2}{q}.
\end{equation*}
Choose $q\geq \frac{2C_2}{\varepsilon}$. Then
\begin{equation}
\label{eq:second-error}
\|\widetilde g_m-F_{m,q}\|_{W^{1,p}(K)}
\leq \frac{\varepsilon}{2}.
\end{equation}
By the triangle inequality, we have 
\[ \|g-F_{m,q}\|_{W^{1,p}(K)}\leq\|g-\widetilde g_m\|_{W^{1,p}(K)}+\|\widetilde g_m-F_{m,q}\|_{W^{1,p}(K)}\leq \varepsilon.\]
It remains to estimate the network size. The max-affine function
$h_m$ can be represented by a convex max-affine network with
$O(m)$ parameters or neurons, while $\psi_q$ has size $O(q)$.
Therefore, $n(F_{m,q})=O(m+q)$.
Since $m=O(\varepsilon^{-d})$ and 
$q=O(\varepsilon^{-1})$
\[n(F_{m,q})=O(\varepsilon^{-d}+\varepsilon^{-1})=O(\varepsilon^{-d}).\]
\end{proof}

\section{Transfer principle for $\relu$-transferable activations}
\label{sec:relutrans}

In this section, we present $\sigma$-ICNN family networks using $\relu$-transferable activations. We show that $\sigma$-ICNN family networks have approximation power and structure-preserving properties similar to those of $\relu$-ICNN family networks.

\begin{theorem}[ICNN Transfer principle for $\relu$-transferable activations]\label{thm:trans_icnn}
Let $\sigma:\mathbb R\to\mathbb R$ be $\relu$-transferable, with associated approximation
$$
\rho_\tau(t)=\alpha_\tau\sigma(\beta_\tau t)+\gamma_\tau \longrightarrow \relu(t).
$$
For given max-affine function 
$$h(x) = \max_{1 \le i \le m } u_i(x), \quad u(x) = a_i^\top x +  b_i,$$ it admits $\sigma$-ICNN approximation $H_\tau$ of width $1$, depth $m-1$, and $O(md)$ parameters such that, for every $1\le p<\infty$,
$ H_\tau\longrightarrow h \;\; \text{in }W^{1,p}(K).$
Likewise, every finite nondecreasing piecewise-linear function $\psi$ admits nondecreasing $\sigma$-network approximation $\psi_\tau$ with the same order of parameter complexity such that
$$
\psi_\tau\longrightarrow\psi
\qquad\text{uniformly and in }W^{1,p}(I),
$$
for every bounded interval $I$ and every $1\le p<\infty$.
\end{theorem}

\begin{proof}
We prove the two assertions separately.
For the max-affine function, write
$$
u_i(x)=a_i^\top x+b_i,
\qquad
h(x)=\max_{1\le i\le m}u_i(x).
$$
Define the $s_i(x)$ as 
$$
s_1(x)=u_1(x),
\qquad s_{k+1}(x)=\max\{s_k(x),u_{k+1}(x)\},\qquad k=1,\ldots,m-1.
$$
Then we have $s_m=h.$
For each $\tau$, define the corresponding smoothed recursion by $s_{1,\tau}(x)=u_1(x),$
and $s_{k+1,\tau}(x)=u_{k+1}(x)+\rho_\tau\bigl(s_{k,\tau}(x)-u_{k+1}(x)\bigr). $ Since $ \rho_\tau(t)= \alpha_\tau\sigma(\beta_\tau t)+\gamma_\tau, $
this recursion is realized by a width-one $\sigma$-ICNN. Indeed, set
$$  z_{1,\tau}(x)  =  \sigma  \left(  \beta_\tau  \bigl(  u_1(x)-u_2(x)  \bigr)  \right).
$$
Then we have 
$$  s_{2,\tau}(x)  =  u_2(x)  +  \alpha_\tau z_{1,\tau}(x)  +  \gamma_\tau.
$$
Inductively, we have 
\begin{align*}
 z_{k+1,\tau}(x)&=\sigma \left( \alpha_\tau\beta_\tau z_{k,\tau}(x) + \beta_\tau \bigl( u_{k+1}(x)-u_{k+2}(x) \bigr) + \beta_\tau\gamma_\tau \right), \\
 s_{k+2,\tau}(x)&=u_{k+2}(x)+\alpha_\tau z_{k+1,\tau}(x)+\gamma_\tau.
\end{align*}
Because $\alpha_\tau\beta_\tau\ge0 $ and $\alpha_\tau>0,$
all hidden-to-hidden and hidden-to-output weights are nonnegative. Thus $H_\tau:=s_{m,\tau}$  is a valid $\sigma$-ICNN of width $1$ and depth $m-1$. Its number of parameters is $O(md)$.

We next prove convergence. Since $K$ is compact and the $u_i$ are affine, all relevant ranges are bounded. Using
$\rho_\tau(t)\longrightarrow \relu(t) $
locally uniformly and the identity
$ v+\relu(u-v)=\max\{u,v\}, $
an induction on $k$ gives
$s_{k,\tau}\longrightarrow s_k$
uniformly on $K$. Hence $H_\tau\longrightarrow h $
uniformly on $K$.
We now consider the gradients. At points of differentiability, we have
$ \nabla s_{k+1,\tau}=\nabla u_{k+1}+\rho_\tau'\bigl(s_{k,\tau}-u_{k+1}\bigr)\bigl(\nabla s_{k,\tau}-\nabla u_{k+1} \bigr).
$

Outside the switching sets of the finite max-affine functions, the comparison between $s_k$ and $u_{k+1}$ is strict. Since $s_{k,\tau}\to s_k$ uniformly and
$\rho_\tau'(t)\longrightarrow\mathbf 1_{\{t>0\}}$
for $t\neq0$, induction gives
$\nabla s_{k,\tau}(x)\longrightarrow\nabla s_k(x)$
for almost every $x\in K$.

The switching sets corresponding to distinct affine pieces are contained in finitely many affine hyperplanes and therefore have Lebesgue measure zero. If two affine pieces coincide identically, their gradients also coincide and cause no difficulty.

Moreover, since
$0\le \rho_\tau'\le C_\sigma, $
the gradient recursion implies, by induction, that
$ \|\nabla s_{k,\tau}(x)\| \le C_k$  for almost every $x\in K$, where $C_k$ is independent of $\tau$. Therefore, by dominated convergence theorem,
$$
\|\nabla H_\tau-\nabla h\|_{L^p(K)}
\longrightarrow0
$$
for every $1\le p<\infty$. Together with the uniform convergence of the function values, this yields
$$
H_\tau\longrightarrow h
\qquad
\text{in }W^{1,p}(K).
$$
We now consider a finite nondecreasing piecewise-linear function $\psi$ on a bounded interval
$I=[a,b]. $
Let
$ a=t_0<t_1<\cdots<t_q=b$ contain all breakpoints of $\psi$, and let $c_j\ge0$ denote the slope of $\psi$ on
$ (t_{j-1},t_j).$
Then, on $I$,
$$
\psi(t) = \psi(a) + \sum_{j=1}^q c_j \left[ \relu(t-t_{j-1}) - \relu(t-t_j) \right].
$$
Define
$$ \psi_\tau(t) = \psi(a) + \sum_{j=1}^q c_j \left[ \rho_\tau(t-t_{j-1}) - \rho_\tau(t-t_j) \right].
$$
Since$
\rho_\tau(t)= \alpha_\tau\sigma(\beta_\tau t)+\gamma_\tau, $
the constants $\gamma_\tau$ cancel in each difference, and $\psi_\tau$ is a two-layer $\sigma$-network with $O(q)$ parameters.
Because $\rho_\tau$ is convex, its derivative is nondecreasing. Thus, for almost every $t$,
$\rho_\tau'(t-t_{j-1})\ge\rho_\tau'(t-t_j),$
since
$ t_{j-1}<t_j. $
It follows that
$$\psi_\tau'(t) = \sum_{j=1}^q c_j \left[ \rho_\tau'(t-t_{j-1}) - \rho_\tau'(t-t_j) \right] \ge 0$$
almost everywhere. Hence $\psi_\tau$ is nondecreasing.
Local uniform convergence of $\rho_\tau$ to $\relu$ gives
$\psi_\tau\longrightarrow\psi $
uniformly on $I$.
Furthermore, for every $t$ outside the finite set of knots,
$$
\rho_\tau'(t-t_j)
\longrightarrow
\mathbf 1_{\{t>t_j\}},
$$
and consequently
$$ \psi_\tau'(t) \longrightarrow \sum_{j=1}^q c_j \left[ \mathbf 1_{\{t>t_{j-1}\}} - \mathbf 1_{\{t>t_j\}} \right] =\psi'(t) $$
for almost every $t\in I$.

Finally, we have 
$|\psi_\tau'(t)|\le2C_\sigma\sum_{j=1}^q c_j, $
uniformly in $\tau$. The dominated convergence theorem therefore gives
$$
\|\psi_\tau'-\psi'\|_{L^p(I)}
\longrightarrow0
$$
for every $1\le p<\infty$. Combining this with the uniform convergence of $\psi_\tau$ yields 
$$
\psi_\tau\longrightarrow\psi
\qquad
\text{uniformly and in }W^{1,p}(I).
$$
The construction uses $O(q)$ parameters, so the order of parameter complexity is unchanged.
This completes the proof.
\end{proof}

\begin{theorem}[IqCNN transfer principle for $\relu$-transferable activations]\label{thm:iqcnn_tranfer}
Let $\sigma:\mathbb R\to\mathbb R$ be $\relu$-transferable, with associated approximation
$$
\rho_\tau(t)=\alpha_\tau\sigma(\beta_\tau t)+\gamma_\tau \longrightarrow \relu(t).
$$
Then the followings hold:
\begin{enumerate}
\item For given  $\relu$-IqCNN, it can be approximated uniformly on $K$ by $\sigma$-IqCNNs preserving quasiconvexity.  
\item For given  $\relu$-IctNN can be approximated in $W^{1,p}(K)$, $1\le p<\infty$, by $\sigma$-IctNNs preserving convex transformability.
\end{enumerate}
In particular, the approximation rates obtained with ReLU remain valid for every $\relu$-transferable activation:
$$
\|f-F\|_{L^\infty(K)}
\le\varepsilon,
\qquad
n(F)=O\left(\varepsilon^{-(d+1)}\right)
$$
for general Lipschitz quasiconvex functions $g \in \Fqcvx(L)$, and
$$
\|f-F\|_{W^{1,p}(K)}
\le\varepsilon,
\qquad
n(F)=O\left(\varepsilon^{-d}\right),
\qquad
1\le p<\infty,
$$
for convex transformable functions $ g \in \Fqcc$.
\end{theorem}

\begin{proof}
We first record a basic consequence of $\relu$-transferability. Let $G$ be a fixed $\relu$-ICNN and $G_\tau$ denote the $\sigma$ network by replacing every occurrence of $\relu$ in $G$ by $\rho_\tau(t)=\alpha_\tau\sigma(\beta_\tau t)+\gamma_\tau.$
Since $\alpha_\tau,\beta_\tau>0$, the factors $\alpha_\tau$ and $\beta_\tau$ can be absorbed into the adjacent weights and $\gamma_\tau$ into the biases without changing the sign of any hidden-to-hidden weight. Hence $G_\tau$, is a $\sigma$-ICNN with the same architecture and the same order of parameter complexity.
Since 
$\rho_\tau\longrightarrow\relu$,
locally uniformly, we have
$G_\tau\longrightarrow G$ uniformly on $K$.
Moreover, for every $1\le p<\infty$,
$G_\tau\longrightarrow G \quad \text{in }W^{1,p}(K).$
Indeed, away from the switching sets of the ReLU units, this follows from
$\rho_\tau'(t)
\longrightarrow
\mathbf 1_{\{t>0\}}. $
On a switching set where a preactivation $v$ vanishes, the identity
$\nabla v=0 \quad\text{a.e. on }\{v=0\}$
shows that the corresponding derivative ambiguity does not contribute to the weak gradient. Since
$ 0\le \rho_\tau'\le C_\sigma,$
the gradients of $G_\tau$ are uniformly bounded on $K$. Dominated convergence theorem therefore gives the claimed $W^{1,p}$ convergence.

Likewise, if $\psi$ is a fixed nondecreasing ReLU network, replacing each ReLU by $\rho_\tau$ gives a nondecreasing $\sigma$-network $\psi_\tau$ satisfying, on every bounded interval $I$,
$ \psi_\tau\longrightarrow\psi$
uniformly and
$$
\psi_\tau\longrightarrow\psi
\qquad\text{in }W^{1,p}(I),
\qquad
1\le p<\infty.
$$
The parameter count remains of the same order.
We now prove the two assertions.
For (1), let a $\relu$-IqCNN be given by
$$
F(x) :=\max_{1\le i\le N} \psi_i\bigl(G_i(x)\bigr),
$$
where each $G_i$ is a ReLU ICNN and each $\psi_i$ is a nondecreasing ReLU network.

Let $G_{i,\tau}$ and $\psi_{i,\tau}$ be their corresponding $\sigma$-approximations, and define
$$F_\tau(x)=\max_{1\le i\le N}\psi_{i,\tau}\bigl(G_{i,\tau}(x)\bigr).$$

Each $G_{i,\tau}$ is convex and each $\psi_{i,\tau}$ is nondecreasing. Therefore,
$\psi_{i,\tau}\circ G_{i,\tau}$
is quasiconvex. Since a finite maximum of quasiconvex functions is quasiconvex, $F_\tau$ is a $\sigma$-IqCNN preserving quasiconvexity.
Because $G_{i,\tau}\longrightarrow G_i$
uniformly on $K$, all ranges $G_{i,\tau}(K)$ are contained, for sufficiently large $\tau$, in a common compact interval $I_i$. Hence
$$
\psi_{i,\tau} \; \circ \; G_{i,\tau}\longrightarrow \psi_i \; \circ \;  G_i
$$
uniformly on $K$. Since the maximum map is $1$-Lipschitz with respect to the sup norm,
$$ \|F_\tau-F\|_{L^\infty(K)} \le \max_{1\le i\le N} \left\| \psi_{i,\tau}\circ G_{i,\tau} - \psi_i\circ G_i \right\|_{L^\infty(K)},
$$
and therefore we have 
$ F_\tau\longrightarrow F\quad\text{uniformly on }K.$
This proves (1).

For (2), let a finite ReLU-IctNN be given by
$F(x)=\psi\bigl(G(x)\bigr),$
where $G$ is a ReLU ICNN and $\psi$ is a finite nondecreasing ReLU network. Let
$F_\tau(x) =\psi_\tau\bigl(G_\tau(x)\bigr).$

Since $G_\tau$ is convex and $\psi_\tau$ is nondecreasing, every $F_\tau$ is convex transformable.
The uniform convergence of $G_\tau$ and $\psi_\tau$ gives
$F_\tau\longrightarrow F$
uniformly on $K$, and hence also in $L^p(K)$.
It remains to prove convergence of the gradients. By the Sobolev chain rule,
$$\nabla F_\tau(x)=\psi_\tau'\bigl(G_\tau(x)\bigr)\nabla G_\tau(x)
$$
for almost every $x\in K$.

Let $T$ denote the finite set of breakpoints of the piecewise-linear function $\psi$. If $G(x)\notin T,$ then the convergence $ G_\tau(x)\to G(x)$
and ReLU-transferability imply
$$ \psi_\tau' \bigl( G_\tau(x) \bigr) \longrightarrow \psi' \bigl( G(x) \bigr). $$
Together with $\nabla G_\tau(x)\longrightarrow\nabla G(x)$
for almost every $x$, we obtain
$\nabla F_\tau(x)\longrightarrow \nabla F(x).$
If instead
$G(x)\in T,$ then, since $G$ is Lipschitz,
$ \nabla G(x)=0 $
for almost every $x$ on the corresponding level set. Since $ \nabla G_\tau(x)\longrightarrow\nabla G(x) $
and $\psi_\tau'$ is uniformly bounded, we again obtain 
$\nabla F_\tau(x)\longrightarrow\nabla F(x)$
for almost every such $x$.

Thus we have $\nabla F_\tau(x)\longrightarrow\nabla F(x)$
for almost every $x\in K$. The derivatives $\psi_\tau'$ and the gradients $\nabla G_\tau$ are uniformly bounded because the networks are finite and $0\le\rho_\tau'\le C_\sigma.$
Hence dominated convergence theorem gives
$$
\|\nabla F_\tau-\nabla F\|_{L^p(K)}
\longrightarrow0,
$$
and therefore
$$
F_\tau\longrightarrow F
\qquad\text{in }W^{1,p}(K),
\qquad
1\le p<\infty.
$$
This proves (2).

Finally, we transfer the approximation rates. For a Lipschitz quasiconvex target $f$, choose a ReLU-IqCNN $F^{\mathrm{ReLU}}$ such that
$\|f-F^{\mathrm{ReLU}}\|_{L^\infty(K)} \le\frac{\varepsilon}{2}, $
with
$n(F^{\mathrm{ReLU}})=O\left(\varepsilon^{-(d+1)} \right).$
By part (1), for sufficiently large $\tau$ there exists a $\sigma$-IqCNN $F_\tau$ with the same order of parameter complexity and
$$
\|F^{\mathrm{ReLU}}-F_\tau\|_{L^\infty(K)}
\le
\frac{\varepsilon}{2}.
$$
Hence $\|f-F_\tau\|_{L^\infty(K)}\le\varepsilon,$ and 
$ n(F_\tau) = O \left( \varepsilon^{-(d+1)} \right).$

Similarly, for a convex transformable target $f$ and $1\le p<\infty$, choose a ReLU-IctNN $F^{\mathrm{ReLU}}$ satisfying
$$
\|f-F^{\mathrm{ReLU}}\|_{W^{1,p}(K)}
\le
\frac{\varepsilon}{2},
$$
with
$ n(F^{\mathrm{ReLU}}) = O \left( \varepsilon^{-d} \right). $
By part (2), for sufficiently large $\tau$ there exists a $\sigma$-IctNN $F_\tau$ such that
$$
\|F^{\mathrm{ReLU}}-F_\tau\|_{W^{1,p}(K)}
\le
\frac{\varepsilon}{2},
$$
with the same order of parameter complexity. Therefore,
$$
\|f-F_\tau\|_{W^{1,p}(K)}
\le
\varepsilon , \quad n(F_\tau)  =  O  \left(  \varepsilon^{-d}  \right).$$
Thus the ReLU approximation exponents are preserved for every $\relu$-transferable activation. 
\end{proof}

\section{Experiments}
\label{sec:exp}

In this section, we evaluate structure-preserving neural architectures for PDE-based approximation of convex, semiconvex, quasiconvex, and convex-transformable functions. We consider four problem classes: (A) convex, (B) semiconvex, (C) quasiconvex, and (D) convex-transformable functions.

Each experiment is formulated through a PDE whose solution belongs to the corresponding function class. We compare structure-preserving architectures, which enforce the relevant geometry by construction, with standard MLPs equipped with sampled shape penalties. The main objective is to assess whether architectural constraints provide more reliable preservation of the target structure than penalties imposed only at finitely many collocation points. We consider problems in three and higher dimensions. Note that quasiconcave functions can be obtained by negation.

\subsection{Setup}

We use ICNN-family architectures and MLPs with Softplus activations so that the required first- and second-order derivatives are well defined. Since Softplus is $\relu$-transferable, the corresponding Softplus-ICNN family retains the structure-preserving and approximation properties established in Theorem~\ref{thm:sigmaapprox}.

The models are trained using physics-informed objectives that combine PDE residuals with the appropriate boundary or auxiliary conditions. The specific PDE and loss terms depend on the function class and are described in the corresponding experiment below. In addition to the PDE loss, penalized MLP baselines include sampled shape constraints tailored to each class.

All results are reported as the test-set mean and standard deviation over three independently initialized runs, and figure error bars indicate one standard deviation. Training collocation points are resampled from the specified domain at every optimization step. We use Adam with cosine learning-rate decay, batch size $768$, and $6{,}000$ fixed test points per run. The semiconvex and convex-transformable experiments are trained for $5{,}000$ steps, while the quasiconvex experiment is trained for $4{,}000$ steps. Architectures are chosen to have comparable numbers of trainable parameters.

For the penalized MLP baselines, the shape penalty is chosen according to the target structure. In part A, we penalize negative values of the minimum Hessian eigenvalue. In part B, we apply the same penalty to the shifted Hessian
$D^2F+\rho_t I.$
In part C, we use the second-order quasiconvexity condition and penalize negative curvature along eight random directions orthogonal to $\nabla F$, corresponding to tangent directions of the level set. In part D, we provide the oracle transform $g=\exp(f)$ and penalize negative eigenvalues of
$D^2\exp(F_{\rm MLP}).$

\begin{figure}[htbp]
\centering
\includegraphics[width=0.96\textwidth]{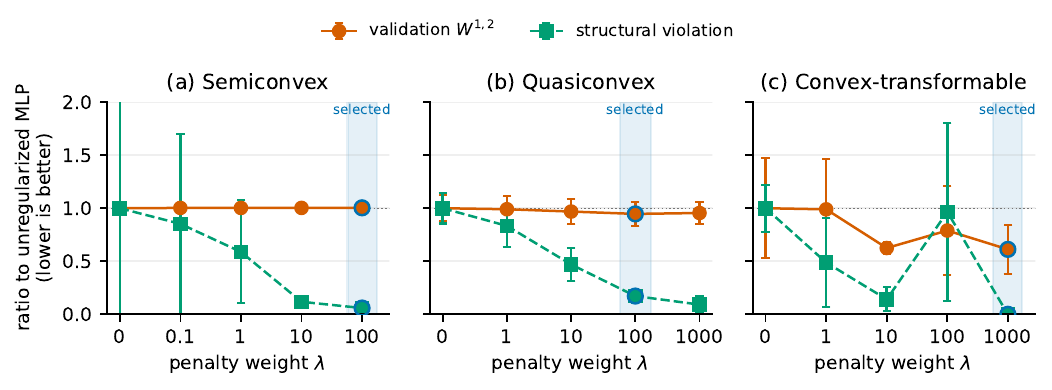}
\caption{Validation selection of the supervised shape penalties in $d=8$.
Each quantity is divided by its value for the unregularized MLP
($\lambda=0$), so lower is better and one means no change.  Curves and error
bars are the mean and one standard deviation from three independently trained
validation runs.  Orange circles show $W^{1,2}$ error, green squares show the
remaining structural violation, and blue rings mark the frozen choices:
$\lambda=100$ in B--C and $\lambda=1000$ in D.}
\label{fig:regularization_pareto}
\end{figure}

Each shape penalty is evaluated on additional unlabeled collocation points. For parts C and D, these points are sampled from the larger evaluation cubes of radii $1.5$ and $2$, respectively.
We tune the penalty parameter on independent validation samples, using
$\lambda\in \{0,0.1,1,10,100\} $
for part B and
$\lambda\in \{0,1,10,100,1000\}$
for parts C and D. Each value of $\lambda$ is evaluated using three independently trained models. Figure~\ref{fig:regularization_pareto} reports the full sweep. Based on validation performance, we fix $\lambda=100$ for parts B and C and $\lambda=1000$ for part D before test evaluation.

These penalized MLPs provide strong finite-sample baselines, but their structural constraints are enforced only at sampled points. In contrast, ICNN, IsCNN, IqCNN, and IctNN satisfy their respective structural constraints globally for every admissible parameter value.

\subsection{(A) Convex target: a Monge–Ampère Dirichlet problem}
\label{sec:exp_convex}
We consider the following Monge–Ampère Dirichlet problem 
\[
 \det D^2u(x)=r(x),\quad x\in K,
 \qquad u=u_\star\quad\text{on } \partial K , \quad K = [-1,1]^d
\]
with Dirichlet boundary condition $u=u_\star$ for $d\in\{2,3,8\}$.  The prescribed solution is
\[
 u_\star(x)=\frac12x^\top Qx
 +\frac{0.22}{2}\log\!\left(\sum_{k=1}^{2d+3}
       \exp(2a_k^\top x+b_k)\right),
 \qquad Q=\operatorname{diag}(0.75,\ldots,1.25),
\]
where $a_k$ are unit random directions and
$r=\det D^2u_\star>0$.  
The interior objective is a stabilized log-determinant
residual, and the Dirichlet boundary loss is given weight $20$.  We use $192$ interior points per step,
$1{,}400$ steps, and approximately $14{,}000$ parameters.  The regularized
MLP additionally uses
$100[0.05-\lambda_{\min}(D^2F)]_+^2$ at the collocation points.

Table~\ref{tab:ma}  reports the approximation errors and the smallest Hessian eigenvalue observed over $512$ independent test points, averaged across runs. ICNN is the only method that simultaneously achieves the smallest Sobolev error and maintains a nonnegative Hessian in every tested dimension. The penalty reduces the magnitude of the MLP's negative curvature, but convexity violations remain at the test points, and the finite-sample penalty provides no global guarantee. This illustrates the practical distinction between architectural convexity preservation and soft regularization.

\begin{table}[htbp]
\centering
\caption{Monge--Amp\`ere test errors and sampled minimum Hessian eigenvalue,
reported as mean (standard deviation) over three runs.  Training uses only PDE
and boundary residuals.}
\label{tab:ma}
\scriptsize
\begin{tabular}{clrrrr}
\toprule
$d$ & model & $L^2$ & grad. $L^2$ & $W^{1,2}$ & $\widehat\lambda_{\min}$ \\
\midrule
2 & ICNN & .176 (.004) & .319 (.010) & .364 (.010) & $+.306$ (.011) \\
  & MLP & .400 (.008) & .921 (.014) & 1.004 (.016) & $-.696$ (.045) \\
  & MLP+convex-reg & .221 (.051) & .745 (.154) & .777 (.162) & $-.121$ (.004) \\
\addlinespace
3 & ICNN & .195 (.014) & .334 (.021) & .386 (.025) & $+.297$ (.023) \\
  & MLP & .676 (.161) & 1.672 (.296) & 1.804 (.335) & $-1.016$ (.130) \\
  & MLP+convex-reg & .428 (.148) & 1.226 (.164) & 1.300 (.204) & $-.251$ (.048) \\
\addlinespace
8 & ICNN & .316 (.024) & .407 (.018) & .515 (.029) & $+.286$ (.012) \\
  & MLP & .671 (.077) & 1.935 (.236) & 2.049 (.247) & $-.767$ (.110) \\
  & MLP+convex-reg & .647 (.044) & 1.860 (.131) & 1.969 (.137) & $-.276$ (.045) \\
\bottomrule
\end{tabular}
\end{table}

\begin{figure}[htbp]
\centering
\includegraphics[width=0.88\textwidth]{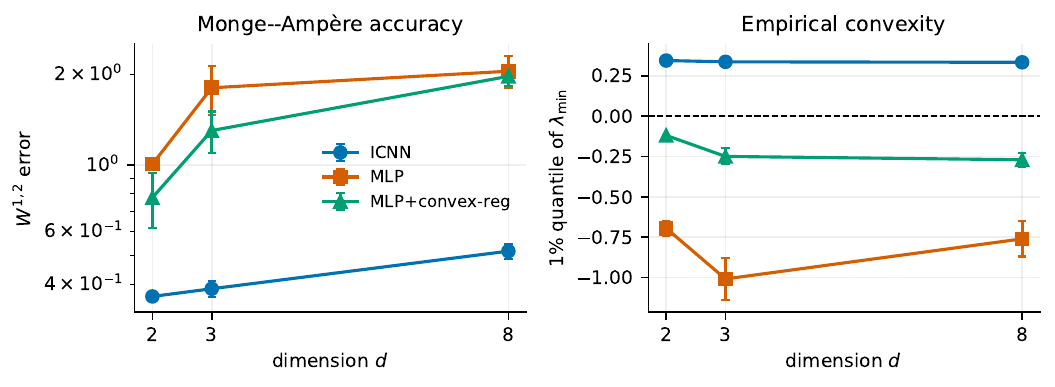}
\caption{Monge--Amp\`ere accuracy (left) and the $1\%$ quantile of the
samplewise minimum Hessian eigenvalue (right).  ICNN remains on the elliptic,
convex branch by construction; a Hessian-regularized MLP reduces but does not
eliminate violations.}
\label{fig:ma}
\end{figure}

\subsection{(B) Semiconvexity target: a Hamilton--Jacobi solution}
\label{sec:exp_semiconvex}

We next consider the Hamilton--Jacobi equation
\[
u_t+\tfrac12\|\nabla_xu\|_2^2=0, \qquad u(0,x)=u_0(x),
\]
whose viscosity solution is given by the Hopf--Lax formula
\[
u(t,x)=\inf_{y\in\mathbb{R}^d}\left\{u_0(y)+\frac{|x-y|^2}{2t} \right\}.
\]
If $u_0(\cdot)$ is $\rho_0$-semiconvex,  for every $0<t<\frac1{\rho_0},$
the function \(u(t,\cdot)\) is \(\rho(t)\)-semiconvex  where $\rho(t) =\frac{\rho_0}{1-\rho_0 t}.$

Hence, we consider the initial condition
\[ u(0,x)=u_0(x):=\sum_{i=1}^d \left(\log\cosh x_i-0.4x_i^2\right).
\]
The function \(u_0\) is \(\rho_0=0.8\)-semiconvex and  nonconvex on a nonempty subset of $[-2,2]^d$.
For \(t=0.35<1/\rho_0=1.25\), the characteristic map
\[ x = y+t\nabla u_0(y) = y+t\bigl(\tanh y-0.8y\bigr)
\]
is invertible. Therefore, the solution can be represented as
\[
u(t,x) = u_0(y) + \frac{t}{2} \left\| \tanh y-0.8y \right\|_2^2,
\]
where $y$ is uniquely determined by
$ x=y+t\bigl(\tanh y-0.8y\bigr)$ and $u(t,x)$ is $\rho$-semiconvex where $\rho_t = \rho_0/(1-t\rho_0)=10/9$. 
We solve the scalar characteristic equation in each coordinate by Newton iteration and use the resulting exact solution snapshot as the target. We consider dimensions $d=2,3,$ and $8$, allowing the same HJ solution to be tested without relying on a spatial grid.

On this smooth supervised problem, the unconstrained MLP achieves slightly lower approximation error than IsCNN (Table~\ref{tab:hj}), but this comes with violations of semiconvexity. In dimensions $2$, $3$, and $8$, its lifted Hessian is indefinite on $8.0\%$,  $20.2\%$, and $2.5\%$ of the $512$ diagnostic samples, respectively. A strong Hessian penalty reduces these rates to $0.33\%$, $0.46\%$, and $0.13\%$, showing that a well-tuned soft constraint can be effective. However, the penalty also increases the low-dimensional approximation error, revealing a tradeoff between accuracy and sampled constraint enforcement rather than providing a global guarantee.

\begin{table}[htbp]
\centering
\caption{Hamilton--Jacobi snapshot errors and fraction of test Hessians that
violate $D^2F+\rho_t I\succeq0$, as mean (standard deviation).}
\label{tab:hj}
\scriptsize
\begin{tabular}{clrrrr}
\toprule
$d$ & model & $L^2$ & grad. $L^2$ & $W^{1,2}$ & violation \\
\midrule
2 & IsCNN & .0072 (.0009) & .0298 (.0042) & .0307 (.0042) & 0 \\
  & MLP & .0025 (.0008) & .0123 (.0033) & .0126 (.0034) & $8.01$ (1.74)\% \\
  & MLP+Hessian-reg & .0080 (.0014) & .0321 (.0048) & .0330 (.0050) & $.33$ (.30)\% \\
3 & IsCNN & .0117 (.0012) & .0476 (.0041) & .0490 (.0042) & 0 \\
  & MLP & .0089 (.0030) & .0352 (.0100) & .0363 (.0104) & $20.18$ (10.16)\% \\
  & MLP+Hessian-reg & .0175 (.0080) & .0592 (.0213) & .0618 (.0226) & $.46$ (.63)\% \\
8 & IsCNN & .184 (.010) & .475 (.025) & .509 (.027) & 0 \\
  & MLP & .164 (.004) & .421 (.012) & .452 (.013) & $2.47$ (2.54)\% \\
  & MLP+Hessian-reg & .165 (.003) & .423 (.011) & .454 (.012) & $.13$ (.23)\% \\
\bottomrule
\end{tabular}
\end{table}

\begin{figure}[htbp]
\centering
\includegraphics[width=0.88\textwidth]{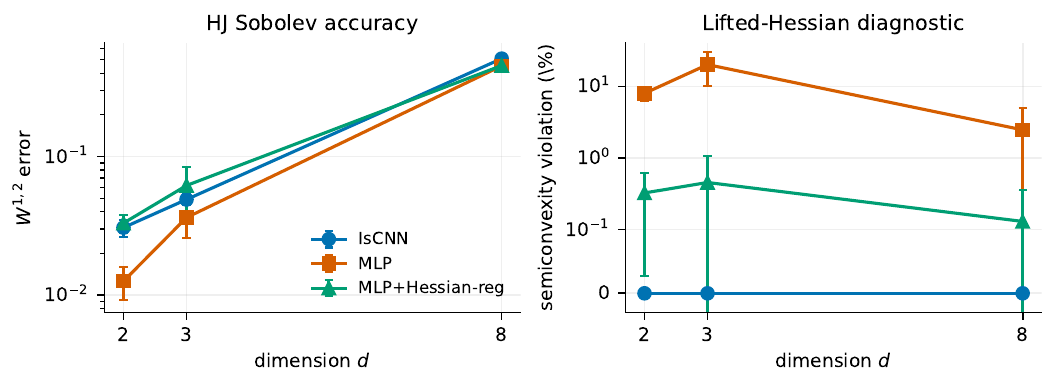}
\caption{HJ Sobolev error (left) and the fraction of sampled lifted Hessians
that violate semiconvexity (right).  IsCNN lies exactly at zero violation;
the regularized MLP approaches zero only on the sampled diagnostic.}
\label{fig:hj}
\end{figure}

\clearpage

\subsection{(C) Quasiconvex target: robust Brascamp--Lieb profiles}
\label{sec:exp_quasiconvex}

We construct three deterministic Brascamp--Lieb-type test functions in dimensions \(d\in\{2,3,8,16\}\). Each function is built from seven rank-one linear projections and a strictly log-concave base function
\[
h(z)=\exp\!\left[-\left(\frac{z^2}{2}+0.06z^4\right)\right].
\]
Because nonnegative weighted products of log-concave functions remain log-concave, 
\[
H_k(x)=\prod_{j=1}^{7} h(b_{kj}^{\top}x-c_{kj})^{s_kp_{kj}}
\]
is log-concave and therefore quasiconcave.
We present the target function 
\[
q(x)=\max_{k=1,2,3}\{\delta_k-H_k(x)\},
\]
which is quasiconvex and combines three intersecting branches. Models are trained on \([-1,1]^d\) and evaluated both in-distribution and on the larger domain \([-1.5,1.5]^d\).
For the MLP baseline, we also use a level-set regularizer that penalizes negative Hessian curvature in randomly sampled directions tangent to level sets. This encourages local quasiconvexity, but it does not provide a global quasiconvexity guarantee.

For completeness, we define the level-set regularizer used for the MLP baseline. At \(n_c=64\) collocation points \(x_i\in[-1.5,1.5]^d\), we sample \(K=8\) independent vectors \(z_{ik}\sim\mathcal N(0,I)\) and construct
\[
 \widehat g_i=\frac{\nabla F(x_i)}
 {\max\{\|\nabla F(x_i)\|_2,\varepsilon\}},\qquad
 v_{ik}=\frac{(I-\widehat g_i\widehat g_i^\top)z_{ik}}
 {\max\{\|(I-\widehat g_i\widehat g_i^\top)z_{ik}\|_2,\varepsilon\}},
\]
with \(\varepsilon=10^{-8}\). The regularization term is
\[
 \mathcal R_{\rm LS}(F)
 =\frac1{n_cK}\sum_{i=1}^{n_c}\sum_{k=1}^K
 \bigl[-v_{ik}^\top D^2F(x_i)v_{ik}\bigr]_+^2.
\]
At points where \(\nabla F(x_i)\neq 0\), the vectors \(v_{ik}\) lie in directions tangent to the corresponding level set. For a smooth quasiconvex function, the Hessian must have nonnegative curvature along such tangent directions. The regularizer therefore penalizes local violations of this necessary condition. Since it is evaluated only at finitely many sample points and directions, however, it does not guarantee global quasiconvexity.

We assess quasiconvexity using two numerical diagnostics. First, we sample \(12{,}000\) triples \((x,y,\lambda)\) and record the fraction for which
\[
F(\lambda x+(1-\lambda)y)>\max\{F(x),F(y)\}.
\]
Second, because uniform random sampling may fail to detect localized violations in higher dimensions, we initialize 128 such triples and optimize them for 60 steps to increase the quasiconvexity violation. We report the normalized maximum violation
\[
 \Delta_{\rm qc}:=
 \frac{\max_{x,y,\lambda}
 [F(\lambda x+(1-\lambda)y)-\max\{F(x),F(y)\}]_+}
 {Q_{0.95}(F)-Q_{0.05}(F)}.
\]
The normalization by the empirical $5\%$--$95\%$ range reduces sensitivity to differences in output scale across models. We also directly assess convexity of empirical sublevel sets by sampling pairs of points from the empirical $25\%$, $50\%$, and $75\%$ sublevel sets and checking whether their midpoints remain in the same sublevel set.

Table~\ref{tab:bl} reports the results on the larger radius-\(1.5\) domain. The unconstrained MLP exhibits a low violation frequency under random sampling, but the optimized violation measure grows substantially with dimension, from \(0.038\) at \(d=2\) to \(0.999\) at \(d=16\). The level-set regularizer provides a stronger baseline: no violations are detected in dimensions \(2\) and \(3\), while the normalized violation is reduced to \(0.070\) and \(0.363\) in dimensions \(8\) and \(16\), respectively. By contrast, IqCNN satisfies the quasiconvexity constraint by construction and therefore has zero violation.

In terms of approximation accuracy, the  $W^{1,2}$ error of IqCNN is $0.193$, nearly the same as that of the unconstrained MLP (\(0.192\)) for $d=16$. The regularized MLP achieves a slightly lower error of \(0.184\), but the numerical tests still identify quasiconvexity violations in higher dimensions.

\begin{table}[htbp]
\centering
\caption{Brascamp--Lieb profile evaluation on the larger cube
$[-1.5,1.5]^d$.  Entries are mean (standard deviation); random failure rates
and $\Delta_{\rm qc}$ measure quasiconvex/sublevel-set violation.}
\label{tab:bl}
\scriptsize
\begin{tabular}{clrrrrr}
\toprule
$d$ & model & $L^2$ & grad. $L^2$ & $W^{1,2}$ & random fail. & $\Delta_{\rm qc}$ \\
\midrule
2 & IqCNN & .0196 (.0099) & .0974 (.0284) & .0994 (.0299) & 0 & 0 \\
  & MLP & .0244 (.0012) & .1168 (.0052) & .1193 (.0053) & $.025$ (.030)\% & .0376 (.0374) \\
  & MLP+levelset-reg & .0199 (.0022) & .1011 (.0069) & .1030 (.0071) & 0 & 0 \\
3 & IqCNN & .0187 (.0092) & .0914 (.0277) & .0933 (.0290) & 0 & 0 \\
  & MLP & .0205 (.0020) & .1017 (.0076) & .1038 (.0078) & $.0028$ (.0048)\% & .0609 (.0160) \\
  & MLP+levelset-reg & .0195 (.0009) & .0939 (.0024) & .0959 (.0025) & 0 & 0 \\
8 & IqCNN & .0611 (.0016) & .1554 (.0040) & .1670 (.0043) & 0 & 0 \\
  & MLP & .0467 (.0077) & .1516 (.0163) & .1587 (.0179) & $.0194$ (.0268)\% & .278 (.034) \\
  & MLP+levelset-reg & .0437 (.0069) & .1438 (.0152) & .1503 (.0165) & $.0028$ (.0048)\% & .0696 (.0601) \\
16 & IqCNN & .0692 (.0005) & .1805 (.0051) & .1933 (.0047) & 0 & 0 \\
   & MLP & .0623 (.0021) & .1819 (.0042) & .1922 (.0046) & $.0139$ (.0048)\% & .999 (.384) \\
   & MLP+levelset-reg & .0588 (.0026) & .1748 (.0039) & .1844 (.0045) & 0 & .363 (.012) \\
\bottomrule
\end{tabular}
\end{table}

\begin{figure}[htbp]
\centering
\includegraphics[width=0.88\textwidth]{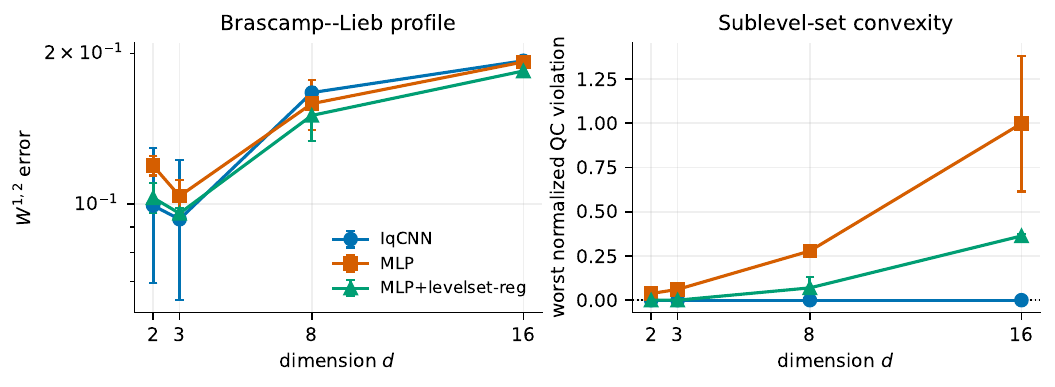}
\caption{Accuracy and adversarial sublevel-set violation on the
Brascamp--Lieb profiles outside the training cube.  The right panel uses an
ordinary linear axis, so exact zeros for IqCNN are displayed at zero.}
\label{fig:bl}
\end{figure}

\clearpage

\subsection{D. Convex-transformable target}
\label{sec:exp_ct}

We construct a high-dimensional convex transformable target $f : \mathbb{R}^d \to \mathbb{R}$: 
\[
g(x)=0.1 +\frac1{40}\sum_{j=1}^{40}w_j
 \frac{\operatorname{Softplus}(3.2a_j^\top x)
 +\operatorname{Softplus}(-3.2a_j^\top x)-2\log2}{3.2} +0.018\sum_{i=1}^d q_i x_i^2,
 \quad f(x)=\log g(x).
\]
Here $w_j>0$, $q_i\in[0.6,1.4]$, and $a_j$ are fixed unit vectors. Since the $\log(\cdot)$ is strictly increasing and $g$ is convex, $f$ is convex transformable. In general, however, $f$ is not convex, because the concavity of the $\log(\cdot)$ can introduce negative curvature.
We construct IctNN with the convex latent function using a shallow ICNN with 80 hidden units and the monotone outer function with a 256-interval calibrator. The MLP widths are chosen so that the parameter counts are comparable: \(1069/1041\), \(1717/1751\), and \(3013/3039\) parameters for IctNN/MLP at \(d=8,16,32\), respectively. All models are trained on the unit cube and evaluated on domains extending to radius \(2\). 

For a stronger comparison with penalty-based methods, we introduce a convex-transform regularization that exploits the true inverse outer transformation. Specifically, it penalizes negative curvature of $D^2\exp(F_{\rm MLP})$
at unlabeled points sampled out to radius \(2\), with the regularization weight fixed at the validation-selected value \(\lambda=1000\).

As reported in Table~\ref{tab:ct}, even with this additional structural information supplied to the MLP, IctNN achieves the lowest in-domain \(W^{1,2}\) error in all three dimensions. At \(d=32\), the errors are \(0.112\), \(0.269\), and \(0.251\) for IctNN, MLP, and the CT-regularized MLP (MLP+CT-reg), respectively. On the radius-\(2\) domain, IctNN remains more accurate in \(d=8\) and has roughly half the error of MLP+CT-reg in \(d=32\). In \(d=16\), however, the regularized MLP gives the lower error, so the comparison does not indicate uniform accuracy dominance. The unconstrained MLP violates convexity of the oracle-transformed output in \(68\)--\(84\%\) of the sampled radius-\(2\) Hessian evaluations. The curvature penalty substantially reduces this frequency, although violations remain in dimensions \(16\) and \(32\). In contrast, IctNN enforces convexity of the learned latent function by construction and restricts the remaining flexibility to a one-dimensional monotone outer transformation.

---------------------------------------------------------------

\begin{table}[htbp]
\centering
\caption{Convex-transformable target: parameter-matched $W^{1,2}$ errors on
the training cube and the larger cube, with the fraction of negative
eigenvalues of $D^2\exp(F)$ at $R=2$.  Entries are mean (standard deviation)
over three independent trainings.  ``latent cert.'' denotes IctNN's global
learned certificate; $\exp$ need not be its learned transform.}
\label{tab:ct}
\scriptsize
\begin{tabular}{clrrrr}
\toprule
$d$ & model & params. & $W^{1,2}_{R=1}$ & $W^{1,2}_{R=2}$ & negative $D^2e^F$ \\
\midrule
8 & IctNN & 1069 & .0458 (.0017) & .165 (.017) & latent cert. \\
  & MLP & 1041 & .312 (.131) & .320 (.148) & $68.10$ (14.42)\% \\
  & MLP+CT-reg & 1041 & .177 (.114) & .193 (.071) & 0 \\
16 & IctNN & 1717 & .0672 (.0017) & .341 (.002) & latent cert. \\
   & MLP & 1751 & .301 (.040) & .375 (.055) & $84.11$ (7.05)\% \\
   & MLP+CT-reg & 1751 & .237 (.047) & .266 (.065) & $.52$ (.90)\% \\
32 & IctNN & 3013 & .1116 (.0006) & .1862 (.0015) & latent cert. \\
   & MLP & 3039 & .269 (.051) & .443 (.100) & $78.13$ (24.24)\% \\
   & MLP+CT-reg & 3039 & .251 (.043) & .376 (.095) & $.46$ (.11)\% \\
\bottomrule
\end{tabular}
\end{table}

\begin{figure}[htbp]
\centering
\includegraphics[width=0.88\textwidth]{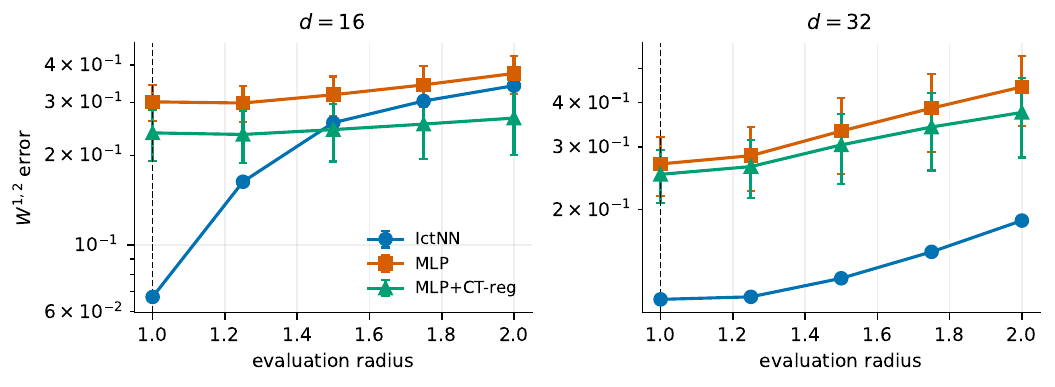}
\caption{Convex-transformable target in dimensions 16 and 32 (mean and one
standard deviation over three independent trainings).  The dashed line is the
training boundary.  IctNN is most accurate in-domain and retains a clear
advantage through radius 2 in $d=32$.}
\label{fig:ct_ood}
\end{figure}

\clearpage

\paragraph{Overall finding.}
The regularized MLPs are useful baselines: they substantially reduce measured
violations and sometimes achieve the lowest error.  They still require
validation and provide only sampled evidence.  The architectural models give
global validity by construction.  In the quasiconvex $d=16$ test IqCNN matches
the plain MLP's Sobolev accuracy while eliminating a large adversarial
violation; in the convex-transformable tests IctNN also gives the lowest
in-domain error through $d=32$.  The evidence supports global structural
control with competitive high-dimensional accuracy, not universal accuracy
dominance.

\section{Conclusion}
\label{sec:conclusion}

We developed a structure-preserving approximation theory for convex, semiconvex, quasiconvex, and convex transformable functions using ICNN-based architectures. The results reveal a common principle: the strength of the approximation guarantee is determined by the form of the underlying convex representation.

For \(C^{1,1}\) convex functions, we established the sharp \(W^{1,p}\) minimax rate \(m^{-1/d}\) under max-affine approximation and showed that every max-affine function can be represented exactly by a bounded-width sequential ICNN with complexity linear in the number of affine pieces. This connects classical piecewise-affine convex approximation directly to neural-network approximation. The same \(W^{1,p}\) approximation order extends to semiconvex functions through quadratic lifting and to convex transformable functions through monotone composition with a convex latent function.

For general Lipschitz quasiconvex functions, the relevant structure is different. Using a supremal representation by monotone responses to affine probes, we derived a quantitative finite-probe approximation theorem and introduced a structure-preserving IqCNN architecture. This yields a value-level \(L^\infty\) approximation guarantee based on a finite family of one-dimensional monotone components. Together, these results distinguish function classes admitting a single recoverable convex latent representation from those described through a family of convexity-compatible probes.

We also showed that the approximation results extend beyond ReLU networks to ReLU-transferable convex activations, including Softplus and ELU, without changing the corresponding approximation orders. This allows the structure-preserving theory to be combined with smooth activations when differentiability of the approximating network is desirable.

The numerical experiments support the theoretical picture. Compared with parameter-matched MLPs and validation-tuned shape penalties, the proposed architectures enforce the prescribed geometry globally by construction, while the penalty-based methods provide only sampled control and require an additional tuning parameter. At the same time, the structure-preserving models retain competitive approximation accuracy, illustrating that exact geometric constraints need not come at the cost of practical approximation quality.

\bibliographystyle{plain} 
\bibliography{ref}

\end{document}